\documentclass[10pt,twocolumn,letterpaper]{article}

\usepackage[pagenumbers]{cvpr} 
\makeatletter
\@namedef{ver@everyshi.sty}{}
\makeatother
\usepackage{graphicx, amsmath, amssymb, caption, subcaption, multirow, overpic, textpos, tabularx}
\usepackage[table]{xcolor}
\usepackage{booktabs}
\usepackage{tikz}
\usepackage{comment}
\usepackage{color}
\usepackage{titletoc}
\usepackage{wrapfig}
\usepackage{pifont}
\usepackage{bm}
\usepackage{amsthm}
\usepackage{makecell}
\usepackage{adjustbox}
\usepackage{wrapfig}

\usepackage[pagebackref,breaklinks,colorlinks]{hyperref}

\newcommand{\shcmt}[1]{\textcolor{blue}{[SH: {#1}]}}

\newcommand{\warn}[1]{\textcolor{green}{#1}}

\newcommand{\cutsubsectiondown}{\vspace*{-0.07in}}
\newcommand{\cutparagraphup}{\vspace*{-0.1in}}

\newlength\savewidth

\newcolumntype{x}[1]{>{\centering\arraybackslash}p{#1pt}}
\newcolumntype{y}[1]{>{\raggedright\arraybackslash}p{#1pt}}
\newcolumntype{z}[1]{>{\raggedleft\arraybackslash}p{#1pt}}

\newcommand{\app}{\raise.17ex\hbox{$\scriptstyle\sim$}}

\definecolor{deemph}{gray}{0.6}

\definecolor{baselinecolor}{gray}{.9}

\usepackage[capitalize]{cleveref}
\crefname{section}{Sec.}{Secs.}
\Crefname{section}{Section}{Sections}
\Crefname{table}{Table}{Tables}
\crefname{table}{Tab.}{Tabs.}

\def\cvprPaperID{9561} 
\def\confName{CVPR}
\def\confYear{2023}

\begin{document}

\title{Towards End-to-End Generative Modeling of Long Videos\\with Memory-Efficient Bidirectional Transformers}
\author{Jaehoon Yoo\textsuperscript{\rm 1}, Semin Kim\textsuperscript{\rm 1},  Doyup Lee\textsuperscript{\rm 2}, Chiheon Kim\textsuperscript{\rm 2}, Seunghoon Hong\textsuperscript{\rm 1}\\
\textsuperscript{\rm 1}KAIST, 
\textsuperscript{\rm 2}Kakao Brain\\
{\tt\small \{wogns98, kimsm1121, seunghoon.hong\}@kaist.ac.kr, \{damien.re, frost.conv\}@kakaobrain.com}}
\maketitle

\begin{abstract}
   Autoregressive transformers have shown remarkable success in video generation. 
    However, the transformers are prohibited from directly learning the long-term dependency in videos due to the quadratic complexity of self-attention, and inherently suffering from slow inference time and error propagation due to the autoregressive process.  
   In this paper, we propose Memory-efficient Bidirectional Transformer (MeBT) for end-to-end learning of long-term dependency in videos and fast inference. 
   Based on recent advances in bidirectional transformers, our method learns to decode the entire spatio-temporal volume of a video in parallel from partially observed patches. 
   The proposed transformer achieves a linear time complexity in both encoding and decoding, by projecting observable context tokens into a fixed number of latent tokens and conditioning them to decode the masked tokens through the cross-attention. 
Empowered by linear complexity and bidirectional modeling, our method demonstrates significant improvement over the autoregressive transformers for generating moderately long videos in both quality and speed. Videos and code are available at \href{https://sites.google.com/view/mebt-cvpr2023}{https://sites.google.com/view/mebt-cvpr2023}.
\end{abstract}

\section{Introduction}


Modeling the generative process of videos is an important yet challenging problem.
Compared to images, generating convincing videos requires not only producing high-quality frames but also maintaining their semantic structures and dynamics coherently over long timescale~\cite{DIGAN, TATS, StyleGAN-V, HVP, CogVideo, CWVAE}.


Recently, autoregressive transformers on discrete representation of videos have shown promising generative modeling performances~\cite{TATS, NUWA, videoGPT, LVT}.
Such methods generally involve two stages, where the video frames are first turned into discrete tokens through vector quantization, and then their sequential dynamics are modeled by an autoregressive transformer.
Powered by the flexibility of discrete distributions and the expressiveness of transformer architecture, these methods demonstrate impressive results in learning and synthesizing high-fidelity videos.


However, autoregressive transformers for videos suffer from critical scaling issues in both training and inference.
During training, due to the quadratic cost of self-attention, the transformers are forced to learn the joint distribution of frames entirely from short videos (\emph{e.g.}, 16 frames~\cite{TATS, videoGPT}) and cannot directly learn the statistical dependencies of frames over long timescales.
During inference, the models are challenged by two issues of autoregressive generative process -- its serial process significantly slows down the inference speed, and perhaps more importantly, autoregressive prediction is prone to error propagation where the prediction error of the frames accumulates over time.

To (partially) address the issues, prior works proposed improved transformers for generative modeling of videos, which are categorized as the following:
\textbf{(a)}~Employing sparse attention to improve scaling during training~\cite{NUWA, MaskViT, CogVideo},
\textbf{(b)}~Hierarchical approaches that employ separate models in different frame rates to generate long videos with a smaller computation budget~\cite{CogVideo, TATS}, and
\textbf{(c)}~Removing autoregression by formulating the generative process as masked token prediction and training a bidirectional transformer~\cite{MaskViT, MMVID}.
While each approach is effective in addressing specific limitations in autoregressive transformers, none of them provides comprehensive solutions to aforementioned problems -- \textbf{(a,~b)}~still inherits the problems in autoregressive inference and cannot leverage the long-term dependency by design due to the local attention window, and \textbf{(c)}~is not appropriate to learn long-range dependency due to the quadratic computation cost.
We believe that an alternative that jointly resolves all the issues would provide a promising approach towards powerful and efficient video generative modeling with transformers.



In this work, we propose an efficient transformer for video synthesis that can fully leverage the long-range dependency of video frames during training, while being able to achieve fast generation and robustness to error propagation.
We achieve the former with a linear complexity architecture that still imposes \emph{dense} dependencies across all timesteps, and the latter by removing autoregressive serialization through masked generation with a bidirectional transformer.
While conceptually simple, we show that efficient dense architecture and masked generation are highly complementary, and when combined together, lead to substantial improvements in modeling longer videos compared to previous works both in training and inference.
The contributions of this paper are as follows:
\begin{itemize}
    \item We propose Memory-efficient Bidirectional Transformer (MeBT) for generative modeling of video.
    Unlike prior methods, MeBT can directly learn long-range dependency from training videos while enjoying fast inference and robustness in error propagation.
    \item To train MeBT for moderately long videos, we propose a simple yet effective curriculum learning that guides the model to learn short- to long-term dependencies gradually over training.
    \item We evaluate MeBT on three challenging real-world video datasets.
    MeBT achieves a performance competitive to state-of-the-arts for short videos of 16 frames, and outperforms all for long videos of 128 frames while being considerably more efficient in memory and computation during training and inference.
\end{itemize}

\section{Background}
%
%
%
This section introduces generative transformers for videos that utilize discrete token representation, which can be categorized into autoregressive and bidirectional models.

Let $\mathbf{x} \in \mathbb{R}^{T \times H \times W \times 3}$ be a video.
To model its generative distribution $p(\mathbf{x})$, prior works on transformers employ discrete latent representation of frames $\mathbf{y} \in \mathbb{R}^{t \times h \times w \times d}$ and model the prior distribution on the latent space $p(\mathbf{y})$.
\cutparagraphup
\paragraph{Vector Quantization}
To map a video $\mathbf{x}$ into discrete tokens $\mathbf{y}$, previous works~\cite{TATS, NUWA, LVT, MMVID, VPVQVAE} utilize vector quantization with an encoder $E$ that maps $\mathbf{x}$ onto a learnable codebook $F=\{e_i\}_{i=1}^U$~\cite{VQVAE}.
Specifically, given a video~$\mathbf{x}$, the encoder produces continuous embeddings $\mathbf{h} = E(\mathbf{x}) \in \mathbb{R}^{t \times h \times w \times d}$ and searches for the nearest code $e_u\in F$.

The encoder $E$ is trained through an autoencoding framework by introducing a decoder $D$ that takes discrete tokens $\mathbf{y}$ and produces reconstruction $\hat{\mathbf{x}}=D(\mathbf{y})$.
The encoder $E$, codebook $F$, and the decoder $D$ are optimized with the following training objective:
\begin{gather}
    \mathcal{L}_q = ||\mathbf{x} - \hat{\mathbf{x}}||_2^2 + ||\texttt{sg}(\mathbf{h}) - \mathbf{y}||_2^2 + \beta ||\texttt{sg}(\mathbf{y}) - \mathbf{h}||^2_2,
\end{gather}
where \texttt{sg} denotes stop-gradient operator.
In practice, to improve the quality of discrete representations, additional perceptual loss and adversarial loss are often introduced~\cite{VQGAN}.

For the choice of the encoder $E$, we utilize 3D convolutional networks that compress a video in both spatial and temporal dimensions following prior works~\cite{videoGPT, TATS}.
\paragraph{Autoregressive Transformers}
Given the discrete latent representations $\mathbf{y} \in \mathbb{R}^{t \times h \times w \times d}$, generative modeling of videos boils down to modeling the prior $p(\mathbf{y})$.
Prior work based on autoregressive transformers employ a sequential factorization $p(\mathbf{y})=\Pi_{i\leq N}p(y_i|y_{<i})$ where $N = thw$, and use a transformer to model the conditional distribution of each token $p(y_i|y_{<i})$.
The transformer is trained to minimize the following negative log-likelihood of training data:
\begin{equation}\label{eq:autoregressive}
    \mathcal{L}_a = \sum_{i\leq N} -\log{p(y_i|y_{<i})}.
\end{equation}

During inference, the transformer generates a video by sequentially sampling each token $y_i$ from the conditional $p(y_i|y_{<i})$ based on context $y_{<i}$.
The sampled tokens $\mathbf{y}$ are then mapped back to a video using the decoder $D$.

While simple and powerful, autoregressive transformers for videos suffer from critical scaling issues.
First, each conditional $p(y_n|y_{<n})$ involves $\mathcal{O}(n^2)$ computational cost due to the quadratic complexity of self-attention.
This forces the model to only utilize short-term context in both training and inference, making it inappropriate in modeling spatio-temporal long-term coherence.
Furthermore, during inference, the sequential decoding requires $N$ model predictions that recursively depend on the previous one.
This leads to a slow inference, and more notably, a potential error propagation over space and time since the prediction error at a certain token accumulates over the remaining decoding steps.
This is particularly problematic for videos since $N$ is often very large as tokenization spans both spatial and temporal dimensions.



\ifdefined\paratitle {\color{blue}
[Limitation1: Quadratic complexity of autoregressive transformers prohibits e2e learning: prior works (sparse attentions) and limitation] \\
} \fi

\begin{figure*}[!ht]
    \centering
    \includegraphics[width=0.9\textwidth]{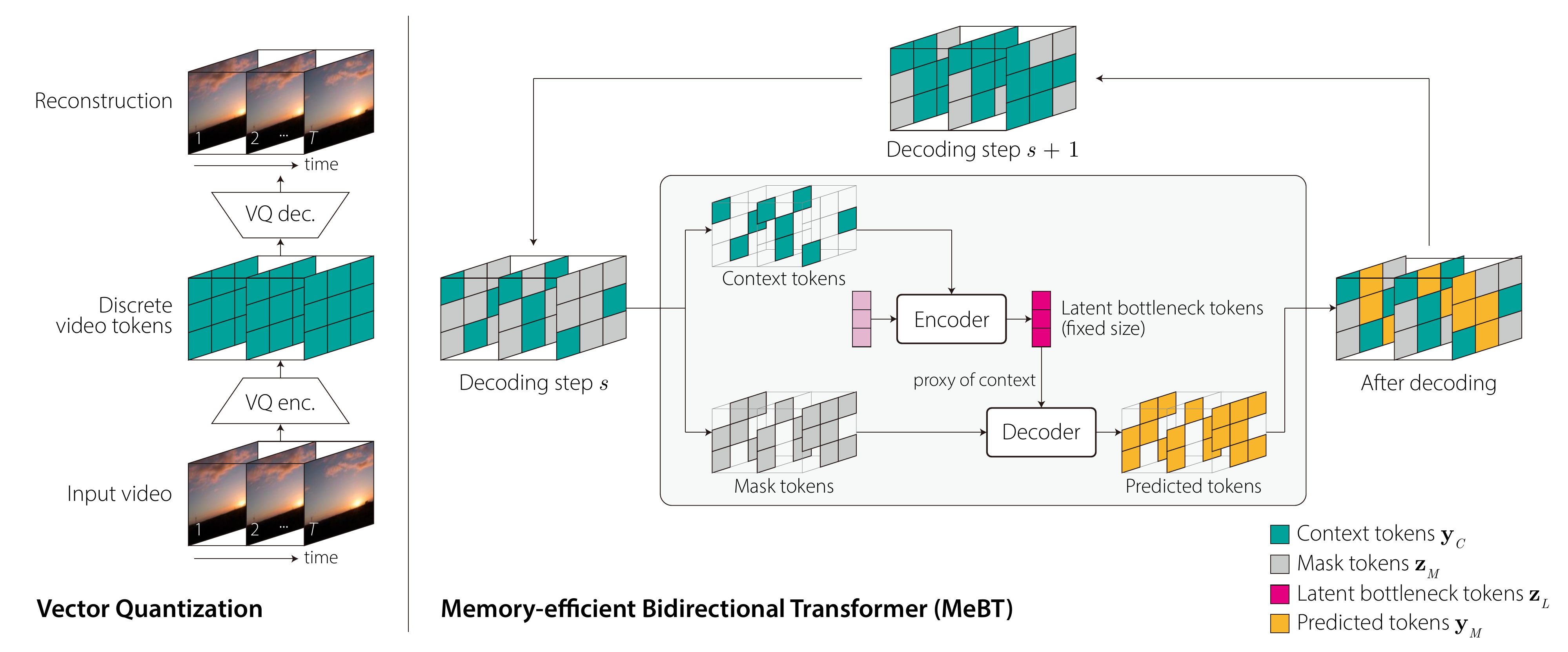}
    \vspace{-0.35cm}
    \caption{Overview of our method. Our model learns to predict masked tokens from the context tokens with linear complexity encoder and decoder. The encoder and decoder utilize latent bottlenecks to achieve linear complexity while performing full dependency modeling. 
    }
    \label{fig:overview}
    \vspace{-0.4cm}
\end{figure*}
\cutparagraphup
\paragraph{Bidirectional Transformers}
\label{sec:maskgit}
\ifdefined\paratitle {\color{blue}
[BT address slow sequential decoding of autoregressive transformers.] \\
} \fi
To improve the decoding efficiency of autoregressive transformers,
bidirectional generative transformers have been proposed~\cite{MaskGIT, M6-UFC, MMVID}.
Contrary to autoregressive models that predict a single consecutive token at each step, a bidirectional transformer learns to predict \emph{multiple} masked tokens at once based on the previously generated context.
Specifically, given the random masking indices $\mathbf{m}\subseteq\{1,...,N\}$, it models the joint distribution over masked tokens $\mathbf{y}_M= \{y_i | i \in \mathbf{m}\}$ conditioned on the visible context $\mathbf{y}_C = \{y_i | i \notin \mathbf{m}\}$, and is trained with the below objective:
\begin{equation}
\label{eq:maskgit}
    \mathcal{L}_b=-\log{p(\mathbf{y}_M|\mathbf{y}_C, \mathbf{z}_M)}
\approx -\sum_{i \in \mathbf{m}} \log{p(y_i|\mathbf{y}_C, \mathbf{z}_M)},
\end{equation}
where mask embeddings $\mathbf{z}_M$ encode positions of the mask tokens with learnable vectors. 
Each conditional in Eq.~\eqref{eq:maskgit} is modeled by a transformer, but contrary to autoregressive models that apply causal masking, bidirectional transformers operate on entire tokens $\mathbf{y}=\mathbf{y}_C\cup \mathbf{y}_M$.
During inference, the model is initialized with empty context (\emph{i.e.}, $\mathbf{y}_C=\emptyset$), and performs iterative decoding by predicting the probability over all masked tokens $\mathbf{y}_M$ by Eq.~\eqref{eq:maskgit} and sampling their subset as the context for the next iteration.


Adopting bidirectional transformers to model videos has two advantages compared to autoregressive transformers.
First, by decoding multiple tokens at once, bidirectional transformers enjoy a better parallelization and smaller decoding steps than autoregressive transformers which results in a faster sampling.
Second, bidirectional transformers are more robust to error propagation than autoregressive counterparts since the decoding of the masked token is independent of the temporal order, allowing consistent prediction quality over time.
However, complexity of the bidirectional transformers is still quadratic to the number of tokens, limiting the length of training videos to only short sequences thus hindering learning long-term dependencies.

\section{Memory-efficient Bidirectional Transformer}

\ifdefined\paratitle {\color{blue}
[To alleviate the problems of prior works, we propose MeBT, a bidirectional transformer with linear complexity that learns to model long-term dependency directly.: slow decoding, error propagation, complexity] \\
} \fi

Our goal is to design a generative transformer for videos that can perform fast inference with a robustness to error propagation, while being able to fully leverage the long-range statistical dependency of video frames in training.

For inference speed and robustness, we adopt the bidirectional approach and parameterize the joint distribution of masked tokens $p(\mathbf{y}_M|\mathbf{y}_C,\mathbf{z}_M)\approx\Pi_{i\in\mathbf{m}}p(y_i|\mathbf{y}_C,\mathbf{z}_M)$ (Eq.~\eqref{eq:maskgit}) with a transformer.
For learning long-range dependency, we take a simple approach of employing an efficient transformer architecture~\cite{EfficientTransformers} of sub-quadratic complexity and directly training it with longer videos.
While sparse attention (\emph{e.g.}, local, axial, strided) is dominant in autoregressive transformers to reduce complexity~\cite{NUWA, MaskViT, CogVideo, TATS}, we find it potentially problematic for bidirectional transformers since the context $\mathbf{y}_C$ can be provided for an arbitrary subset of token positions that often cannot be covered by the fixed sparse attention patterns\footnote{Note that, in autoregressive transformers, the context $\mathbf{y}_{<i}$ (Eq.~\eqref{eq:autoregressive}) is always the entire past, allowing sparse attention to robustly see the context.}.
Thus, unlike prior work, we design an efficient bidirectional transformer based on low-rank latent bottleneck~\cite{SetTransformer, Linformer, SetVAE, Perceiver, PerceiverIO, LUNA} that always enables a \emph{dense} attention dependency over all token positions while guaranteeing \emph{linear} complexity.


To this end, we propose Memory-efficient Bidirectional Transformer (MeBT) for generative modeling of videos.
The overall framework of MeBT is illustrated in Fig.~\ref{fig:overview}.
To predict the masked tokens $\mathbf{y}_M$ from the context tokens $\mathbf{y}_C$, MeBT employs an encoder-decoder architecture based on a fixed number of latent bottleneck tokens $\mathbf{z}_L\in\mathbb{R}^{N_L\times d}$ with $N_L \ll N$.
The encoder projects the context tokens $\mathbf{y}_C$ to the latent tokens $\mathbf{z}_L$, and the decoder utilizes the latent tokens $\mathbf{z}_L$ to predict the masked tokens $\mathbf{y}_M$.
Overall, the encoder and decoder exploit the latent bottleneck to achieve a linear complexity $\mathcal{O}(N)$ while performing a full dependency modeling across the token positions.
In the following sections, we describe the details of the encoder, decoder, and the training and inference procedures of MeBT.

\subsection{Encoder Architecture}
\label{sec:mebt_encoder}
\begin{figure}[!th]
    \centering
    \includegraphics[width=0.47\textwidth]{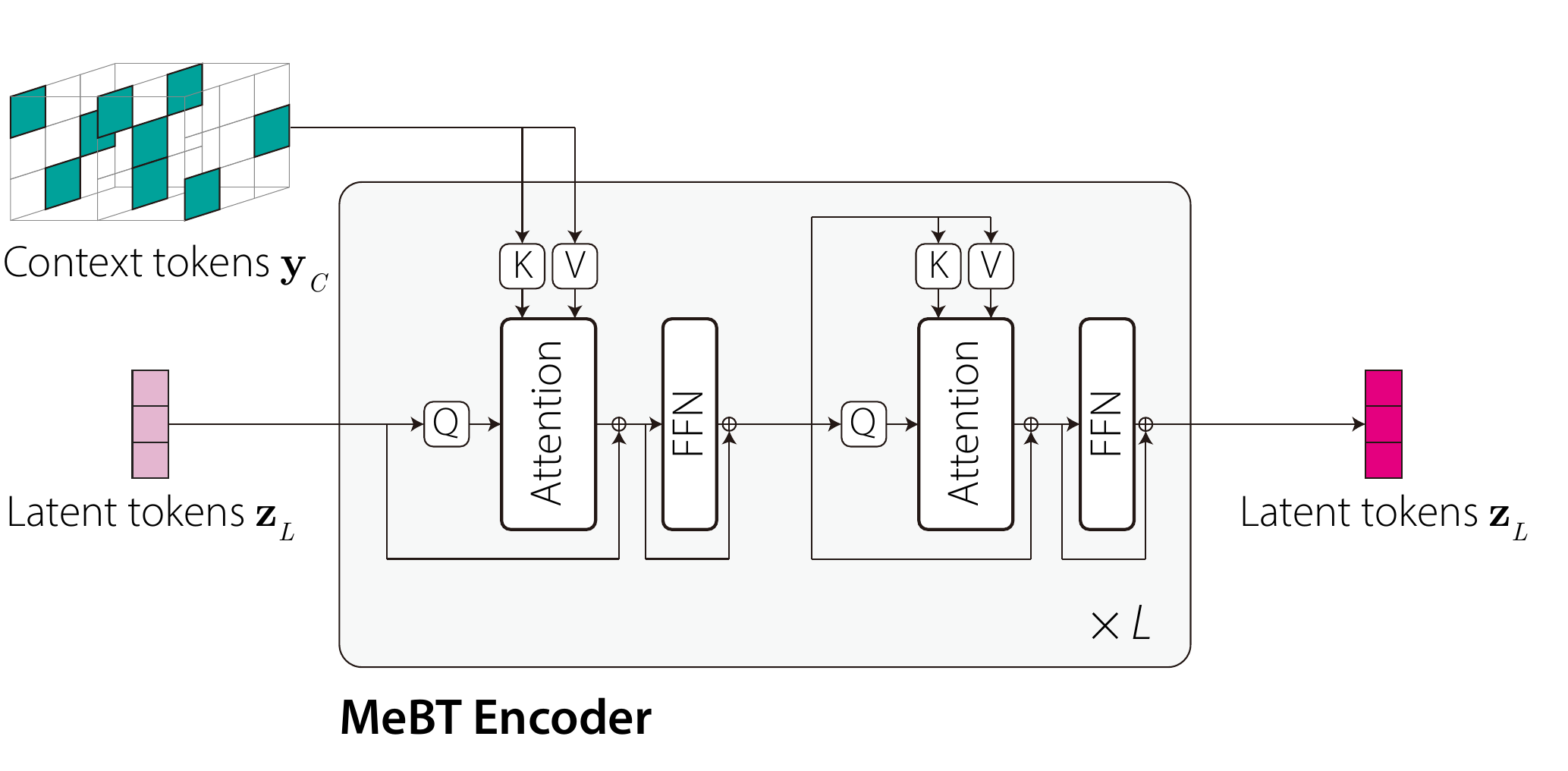}
    \vspace{-0.3cm}
    \caption{Encoder architecture. Our encoder compresses the context tokens into fixed number of latent tokens.}
    \label{fig:encoder}
    \vspace{-0.5cm}
\end{figure}
The encoder aims to project all previously generated context tokens $\mathbf{y}_C$ to fixed-size latent bottleneck $\mathbf{z}_L$ with a memory and time cost linear to context size $\mathcal{O}(N_C)$.
Following prior work on transformers with latent bottlenecks~\cite{Perceiver, PerceiverIO, SetTransformer, LUNA, SetVAE}, we construct the encoder as an alternating stack of two types of attention layers that progressively update the latent tokens based on the provided context tokens.
Specifically, as illustrated in Fig.~\ref{fig:encoder}, the first layer updates the latent tokens by cross-attending to context tokens, and the second one updates the latent tokens by performing self-attention within them.
By stacking the layers, the encoder progressively gathers appropriate contextual information into the latent tokens as well as flexibly updating their state through the exchanges of latent information.

Our encoder with latent bottleneck can benefit bidirectional modeling of videos in several ways.
First, since we employ the fixed-size latent tokens independent to the video length, the complexity of the encoder scales linearly to the number of context tokens, allowing it to directly learn and model long-range dependency by taking longer videos with a reduced memory footprint.
More importantly, unlike the sparse attention~\cite{Ho2019, MaskViT} or hierarchical methods~\cite{CogVideo, TATS} that capture partial dependency among the predefined local subset of context tokens, our encoder always captures full observations of any given set of context tokens by exploiting the latent bottleneck.
This property suits well with the bidirectional modeling since the context tokens can be given as an arbitrary subset of a video often beyond the hand-designed patterns of sparse attention. 

\subsection{Decoder Architecture}
\label{sec:mebt_decoder}
Given the latent bottleneck $\mathbf{z}_L$ as a proxy of context\footnote{
Since the latent tokens $\mathbf{z}_L$ are a proxy of context, we use \emph{latent} and \emph{context} interchangeably to refer to them when describing the decoder.}, the decoder aims to predict the masked tokens $\mathbf{y}_M$ by updating the mask embeddings $\mathbf{z}_M$.
Since the standard masked modeling~\cite{MaskGIT, BERT, MAE} of applying a bidirectional transformer on mask and context $\{\mathbf{z}_M, \mathbf{z}_L\}$ leads to a complexity of $\mathcal{O}((N_M+N_L)^2)$, we opt into an efficient decoder that reduces the cost to linear to mask size $\mathcal{O}(N_M)$ while preserving as much modeling capacity as possible.

\begin{figure}[t]
    \centering
    \includegraphics[width=0.47\textwidth]{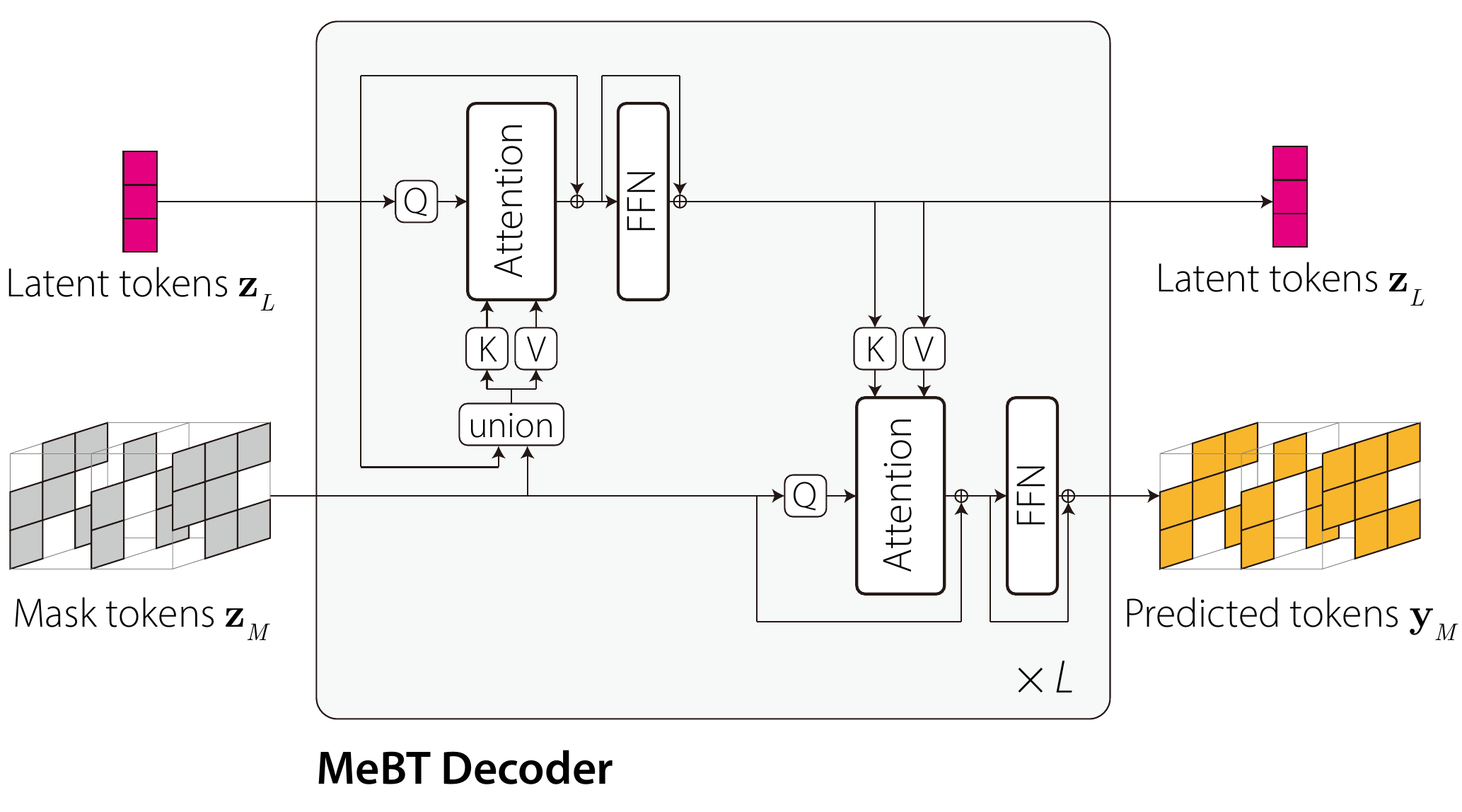}
    \vspace{-0.3cm}
    \caption{Decoder architecture. Our decoder predicts the masked tokens based on the latent tokens encoding the entire context.}
    \label{fig:decoder}
    \vspace{-0.4cm}
\end{figure}

We construct our decoder as a stack of two types of attention layers that update the latent tokens $\mathbf{z}_L$ and mask tokens $\mathbf{z}_M$ in an alternating manner.
More specifically, as illustrated in Fig.~\ref{fig:decoder}, the first layer updates latent tokens $\mathbf{z}_L$ by attending to \emph{both} latent and mask $\{\mathbf{z}_L, \mathbf{z}_M\}$, and the second layer updates mask tokens $\mathbf{z}_M$ by attending to the latent tokens $\mathbf{z}_L$.
After being processed by the decoder, the mask tokens $\mathbf{z}_M$ are then used to predict the masked tokens $\mathbf{y}_M$.

Our decoder achieves linear complexity while retaining a high expressive power, thus being suitable for directly learning and generating long-term dependency.
Since the two attention types have $\mathcal{O}(N_L \times (N_L+N_M)))$ and $\mathcal{O}(N_LN_M))$ complexities respectively, our decoder overall achieves a linear complexity to mask size $N_M$.
Still, we note that this efficient decoder retains much of the modeling capacity of standard masked decoders~\cite{BERT, MAE, MaskGIT} that employ full self-attention on context and mask $\{\mathbf{z}_L, \mathbf{z}_M\}$.
This is seen by decomposing their \emph{all-to-all} dependency $\{\mathbf{z}_L, \mathbf{z}_M\}\to\{\mathbf{z}_L, \mathbf{z}_M\}$ into \emph{all-to-context} $\{\mathbf{z}_L, \mathbf{z}_M\}\to \mathbf{z}_L$, \emph{context-to-mask} $\mathbf{z}_L\to\mathbf{z}_M$, and \emph{mask-to-mask} $\mathbf{z}_M\to\mathbf{z}_M$ dependencies.
In our efficient decoder, the all-to-context and context-to-mask dependencies are precisely modeled by the two types of attention layers, and the mask-to-mask dependency is approximately modeled through the \emph{composition} of the layers by exploiting the latent tokens $\mathbf{z}_L$ as an intermediate bottleneck.
More specifically, to model $\mathbf{z}_M\to\mathbf{z}_M$, our decoder imposes a low-rank approximation through the fixed-size latent representation $\mathbf{z}_M\to\mathbf{z}_L\to\mathbf{z}_M$.
While such approximation appears in a range of prior work on efficient transformers~\cite{SetTransformer, Linformer, LUNA, SetVAE}, to our knowledge, we are the first to employ it for bidirectional video generation.

\subsection{Training and Inference}
\label{sec:train_inference}
\paragraph{Training} 
To train bidirectional transformers for masked token prediction objective in Eq.~\eqref{eq:maskgit}, prior works~\cite{MaskGIT,MaskViT} employed random sampling of the masking indices $\mathbf{m}$. 
Specifically, it choose masking ratio $r \in (0, 1]$ according to a predefined schedule function, and sample $N_M = \left \lceil r N \right \rceil$ unique masking indices randomly by $\mathbf{m}\sim\text{Uniform}(1,N)$.
Although such strategy works well with images or short videos, we observe that training with the strategy directly on moderately long videos often leads to suboptimal solution.  
This is presumably because sampling temporally distant tokens at early stage of training makes the task too difficult, preventing model to learn important local structures. 

\begin{figure}
    \centering
    \includegraphics[width=0.45\textwidth]{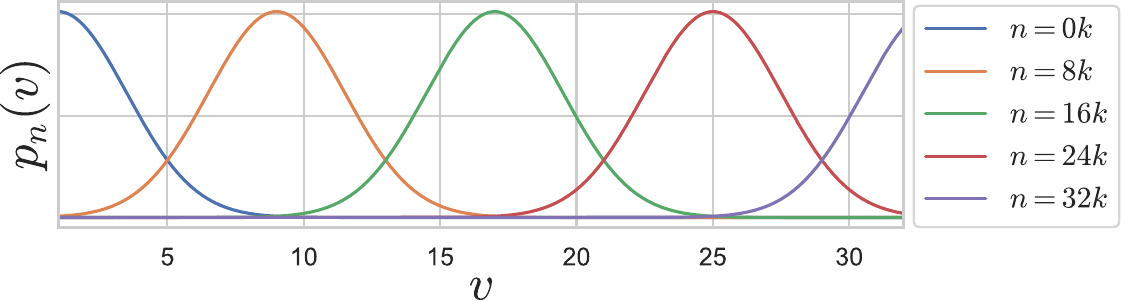}
    \vspace{-0.2cm}
    \caption{Visualization of the interval scheduling function. 
    }
    \label{fig:video_schedule}
    \vspace{-0.7cm}
\end{figure}

To resolve the issue, we propose a simple curriculum-based training that guides the model to learn from short- to long-term sequences gradually over training. 
To this end, we define an \emph{interval} scheduling function that dictates the lengths (intervals) $v$ of training videos at a particular training iteration $n$ by:
\begin{equation}
    p_n(v) \propto \text{exp}\left( - \frac{(v-1 - \frac{n}{\alpha})^2}{2\beta^2} \right)
    \label{eq:masking_schedule}
\end{equation}
where $\alpha$ and $\beta$ are hyperparameters. 
Eq.~\eqref{eq:masking_schedule} characterizes a Gaussian distribution of training video lengths $v$ whose mean gradually increases by the training iterations $n$ with a speed determined by $\alpha$ (Fig.~\ref{fig:video_schedule}).
At every training iteration $n$, we sample the interval $v\in(1,t]$ by first sampling $\hat{v}\sim p_n(v)$ and truncating it within a valid range by $v=\min(\max(1,\lceil\hat{v}\rceil),t)$. 
Then we sample $v$ consecutive frames randomly in a video, and sample the masking indices $\mathbf{m}$ and the context tokens randomly within the sampled interval as discussed above.
We found that such curriculum learning is effective in stabilizing training and improving the prediction quality for long videos (Section~\ref{sec:ablation_study}).

\vspace{-0.2cm}
\paragraph{Inference} 
During inference, MeBT generates videos by iteratively decoding subsets of mask tokens.
To this end, we define a mask sampling function $\gamma$ on $[0, 1]$ that decreases from $\gamma(0) = 1$ to $\gamma(1) = 0$.
At each decoding step $s \in \{0, \cdots, S-1\}$, our method produces predictions on the mask tokens by Eq.~\eqref{eq:maskgit}, and updates the context tokens with top $N_s$ mask tokens according to their sampled probability where $N_s = N - \left \lceil \gamma(\frac{s}{S}) N \right \rceil$.
The process is iteratively applied $S$ times until all tokens are included in the context.
In addition to the na\"ive decoding, we also adopt the revision phase proposed in \cite{draft_and_revise}. 
In this stage, we randomly divide the indices evenly into $R$ partitions $\Pi = \{\mathbf{m}_1, \cdots, \mathbf{m}_R\}$, and revise the tokens in $\mathbf{m}_i$ based on the contexts. 
We found that the additional revision is helpful to improve the consistency and fidelity of generated videos.




\section{Related Work}

\paragraph{Video Generative Models}

Video generative models have shown rapid growth with the advance of generative frameworks.
Various approaches based on Generative Adversarial Networks (GAN)~\cite{DVDGAN,TGAN,TGANv2,MoCoGAN,MoCoGAN-HD,LDVDGAN}, Variational Autoencoders (VAE)~\cite{SVG, SAVP, FitVid, CWVAE, SRVP}, and Implicit Neural Representations (INR)~\cite{StyleGAN-V, DIGAN} are proposed to extend the image generative models to powerful video sequence models.
Despite the impressive success, these methods are mostly displayed in simple and short videos, while progress in modeling complex scenes and motions is still behind ones in the image domains.
Recently, autoregressive transformers~\cite{LVT, videoGPT, TATS, NUWA, CogVideo} demonstrated promising results in modeling complex videos, closing the gap between the image and video generative models.
However, they are mostly limited in short-term videos due to quadratic computation, and inherently suffer from slow inference time and error propagation due to autoregressive process.
This motivates us to seek efficient yet robust transformers for long-term videos.
Long-term video generation has been studied under various context, often conditioned on powerful side information such as real frames~\cite{CWVAE, SRVP, FitVid, LeeKCKR21, GHVAE, ImprovedVRNN}, dense pixel labels~\cite{HVP, Akan2022, LiangLDX17}, action sequences~\cite{Villegas2017, KimNCK19}, or text description~\cite{CogVideo,NUWA}.
Apart from these, we study the potential of transformers for unconditional long-term video generation.  
\cutparagraphup
\vspace{-0.2in}
\paragraph{Efficient Transformers}
\label{sec:effi_transformers}
To improve the scalability of transformers for long sequences, a large volume of work has been developed to approximate the quadratic computation. 
One promising direction is sparse attention that approximates dense pair-wise attention by restricting the size of the neighborhood that each token can attend to~\cite{MaskViT, NUWA, CogVideo, BigBird, Ho2019, LongFormer}. 
However, the fixed sparsity patterns are often incomprehensive to cover various token-wise dependencies in broad applications, especially when combined with masked bidirectional transformers with high masking ratio.
The other line of work proposed latent bottleneck \cite{SetTransformer, SetVAE, LUNA, Perceiver, PerceiverIO} that projects the input to fewer latent embeddings. 
Latent bottleneck can be easily combined with bidirectional transformers as it models dense dependencies between tokens, and this work is the first attempt to demonstrate it in videos. 
\begin{table*}[!ht]
\centering\small
\caption{Quantitative results on 16-frame video generation. * denotes the models trained on both training and validation splits.} 
\vspace{-0.2cm}
\label{tab:short-quantitative}
\begin{subtable}[h]{.56\textwidth}
\centering\small
\caption{SkyTimelapse} 
\begin{tabular}{lcccc}
    \toprule
    Method     & Memory (GB) & Inf.Time & FVD$_{16}$ $(\downarrow)$  & KVD$_{16}$ $(\downarrow)$ \\
    \midrule
    MoCoGAN-HD & 22.4 & {0.04s} & 183.6\tiny{$\pm 5.2$} & 13.9\tiny{$\pm 0.7$} \\
    DIGAN & 4.8 & {0.04s} & 114.6\tiny{$\pm 4.9$} & 6.8\tiny{$\pm 0.5$} \\ 
    TATS-base & 21.9 & 14.3s & 132.6\tiny{$\pm 2.6$} & {5.7}\tiny{$\pm 0.3$} \\ \hline
    MeBT (Ours) & 9.5 & 1.10s & \bf{90.4\tiny{$\pm1.8$}} & \bf{2.6\tiny{$\pm0.1$}} \\
    \bottomrule
\end{tabular}
\vspace{0.1cm}
\caption{Taichi-HD} 
\begin{tabular}{lcccc}
    \toprule
    Method     & Memory (GB) & Inf.Time & FVD$_{16}$ $(\downarrow)$  & KVD$_{16}$ $(\downarrow)$ \\
    \midrule
    MoCoGAN-HD & 22.4 & {0.04s} & 144.7\tiny{$\pm 6.0$} & 25.4\tiny{$\pm 1.9$} \\
    DIGAN & 4.8 & {0.04s} & 128.1\tiny{$\pm 4.9$} & 20.6\tiny{$\pm 1.1$} \\ 
    TATS-base & 21.9 & 14.3s & \bf{94.6}\tiny{$\pm 2.7$} & \bf{9.8}\tiny{$\pm 1.0$} \\ \hline
    MeBT (Ours) & 9.5 & 1.36s & 122.7\tiny{$\pm 4.9$} & 16.9\tiny{$\pm 1.11$} \\
    \bottomrule
\end{tabular}
\end{subtable}
\hfill
\begin{subtable}[h]{0.43\textwidth}
\centering\small
\caption{UCF-101}\label{tab:main_ucf} 
\begin{tabular}{lcc}
\toprule
Method & IS $(\uparrow)$  & FVD$_{16}$  $(\downarrow)$  \\
\midrule
TGAN  & 11.85\tiny{$\pm .07$}   & - \\
LDVD-GAN & 22.91\tiny{$\pm .19$}   & - \\
VideoGPT & 24.69\tiny{$\pm .30$}   & - \\
MoCoGAN-HD*  & 32.36 & 838 \\
DIGAN & 29.71\tiny{$\pm .53$}   & 655\tiny{$\pm 22$}   \\
DIGAN*  & 32.70\tiny{$\pm .35$}   & 577\tiny{$\pm 21$}   \\
CCVS*+StyleGAN  & 24.47\tiny{$\pm .13$} & \bf{386}\tiny{$\pm 15$} \\
StyleGAN-V*  & 23.94\tiny{$\pm .73$} & - \\
Video Diffusion*  & 57.00\tiny{$\pm .62$} & - \\
TATS-base & 57.63\tiny{$\pm .24$} & 420\tiny{$\pm 18$} \\ \hline
Real data & 90.52 & - \\ \hline
MeBT (Ours) & \bf{65.93}\tiny{$\pm .79$} & 438\tiny{$\pm 18$}\\
\bottomrule
\end{tabular}
\end{subtable}
\vspace{-0.3cm}
\end{table*}

\section{Experiments}
\label{sec:experiments}

\subsection{Experimental Setup}
\cutsubsectiondown
\paragraph{Datasets}
We evaluate our method on three popular video benchmarks; SkyTimelapse \cite{stl}, Taichi-HD \cite{taichi}, and UCF-101 \cite{ucf}.
For each dataset, we use 16- and 128-frame-long videos for evaluation of short-term and long-term generation, respectively.
Following prior works, all videos are pre-processed with $128\times128$ spatial resolution.
More details are described in the Appendix (Section \ref{appx:datasets}).

\cutparagraphup
\paragraph{Evaluation metrics}
We evaluate the quality of the generated videos using Fréchet Video Distance (FVD) \cite{FVD} and Kernel Video Distance (KVD) \cite{FVD} based on an I3D network trained on Kinetics-600 \cite{kinetics}. 
For 16-frame videos, we follow the evaluation protocol of \cite{TGANv2} and report the average score over 10 runs where each run samples 2048 videos.
For 128-frame videos, we sample 512 videos 5 times and report the average score.
For UCF-101 dataset, we also measure the Inception Score (IS)~\cite{IS} with a pre-trained C3D model~\cite{c3d} following the prior works~\cite{TATS, CCVS, VGAN}. 
To evaluate the generation quality over time on longer videos (128 frames), we also measure FVD for every 8 frames using a sliding window of 16 frame length.
Furthermore, we evaluate the efficiency of the models by measuring the peak training memory and inference time with batch size 4.
\begin{figure*}[!ht]
    \begin{minipage}{\textwidth}
    \captionsetup{type=table}
    \centering
    \caption{Quantitative results on 128-frame video generation. The subscripts on methods denote the length of training videos.}
    \vspace{-0.2cm}
    \begin{adjustbox}{width=1.0\textwidth}
    \label{tab:longterm}
    \centering
        \begin{tabular}{lccccccccc}
        \toprule
        \\[-1em]& \multicolumn{3}{c}{SkyTimelapse} & \multicolumn{3}{c}{Taichi-HD} & \multicolumn{3}{c}{UCF-101} \\
        \cmidrule(lr){2-4} \cmidrule(lr){5-7} \cmidrule(lr){8-10}
        Method 
        & FVD$_{128}$  $(\downarrow)$ & KVD$_{128}$  $(\downarrow)$ & Inf.Time & FVD$_{128}$  $(\downarrow)$ & KVD$_{128}$  $(\downarrow)$ & Inf.Time
        & FVD$_{128}$  $(\downarrow)$ & KVD$_{128}$  $(\downarrow)$ & Inf.Time\\
        \\[-1em]\Xhline{2\arrayrulewidth}
        \\[-1em]MoCoGAN-HD$_{128}$ & 498\tiny{$\pm 15$} & 23\tiny{$\pm1.39$} & 0.37s & 991\tiny{$\pm23$} & 290\tiny{$\pm12$} & 0.37s & 1622\tiny{$\pm53$} & 118\tiny{$\pm6.4$} & 0.37s\\
        \\[-1em]DIGAN$_{128}$ & 331\tiny{$\pm13$} & 9\tiny{$\pm0.38$} & {0.12s} & 764\tiny{$\pm19$} & 230\tiny{$\pm25$} & {0.12s} & 1627\tiny{$\pm51$} & 122\tiny{$\pm10$} & {0.12s}\\\hline
        \\[-1em]CCVS$_{16}$ & N/A & N/A & N/A & N/A & N/A & N/A & 1411\tiny{$\pm45$} & 121\tiny{$\pm11$} & 219s \\
        \\[-1em]TATS-base$_{16}$ & 435\tiny{$\pm12$} & 19\tiny{$\pm1.50$} & 110s & 458\tiny{$\pm21$} & 211\tiny{$\pm18$} & 110s & 1107\tiny{$\pm31$} & 91.8\tiny{$\pm8.5$} & 110s\\
        \\[-1em]TATS-hierarchical$_{128}$ & 455\tiny{$\pm12$} & 21\tiny{$\pm0.99$} & 127s & 803\tiny{$\pm40$} & 371\tiny{$\pm31$} & 127s & 1138\tiny{$\pm53$} & 83.4\tiny{$\pm$11.6} & 127s\\
        \\[-1em]MeBT$_{128}$ (Ours) & \bf{239\tiny{$\pm3.6$}} & \bf{5.1\tiny{$\pm0.16$}} & 6.53s & \bf{399\tiny{$\pm19$}} & \bf{139\tiny{$\pm13$}} & 6.62s & \bf{968\tiny{$\pm75$}} & \bf{75.5\tiny{$\pm12.1$}} & 7.53s\\
        \bottomrule
        \end{tabular}
    \end{adjustbox}
    \end{minipage}
    \vspace{0.1cm}
    
    \begin{minipage}{\textwidth}
    \centering
    \includegraphics[width=1.0\textwidth]{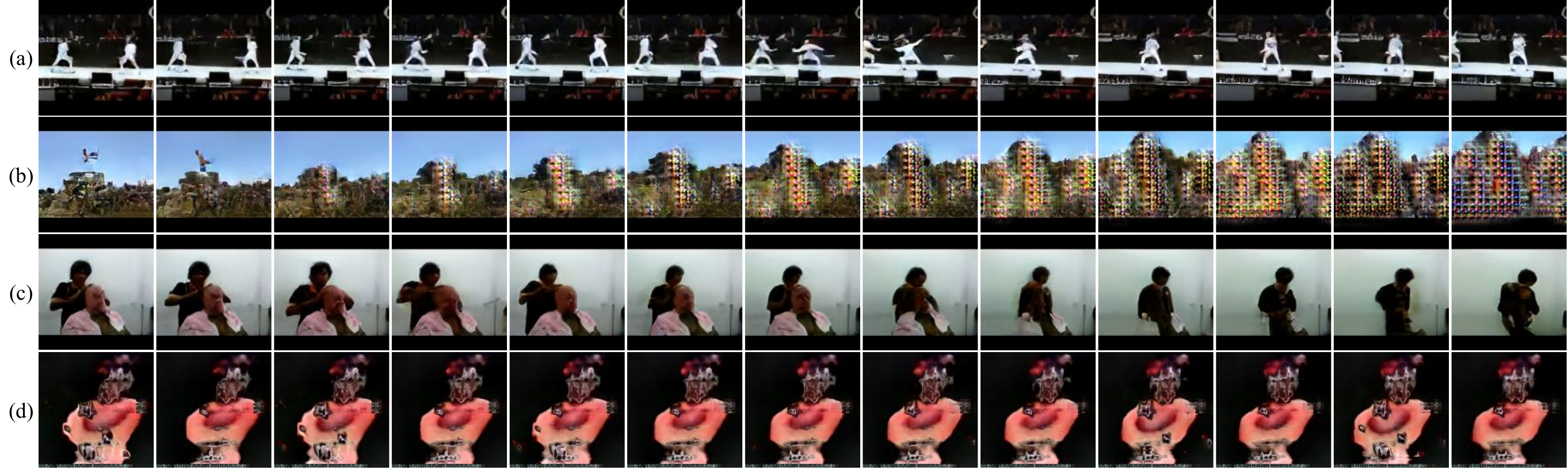} 
    \vspace{-0.5cm}
    \captionsetup{type=figure}
    \vspace{-0.2cm}
    \caption{Qualitative results on 128-frame video generation with different models on the UCF-101 dataset: (a) MeBT (Ours), (b) TATS-base, (c) TATS-hierarchical, (d) DIGAN. 
    We present every 10th frames from the generated videos.
    More results are in Appendix (Fig.~\ref{fig:supp_stl}-\ref{fig:supp_ucf}).
    }
    \label{fig:longterm_qualitative}
    \end{minipage}
    \vspace{-0.6cm}
\end{figure*}
\vspace{-0.4cm}
\paragraph{Implementation}
For discrete tokenization, we use the official VQGAN checkpoints from \cite{TATS} that have a compression ratio of 4 and 8 in temporal and spatial dimensions, respectively.
For a fair comparison, we match the number of parameters and layers in our method to be roughly the same as other transformer-based baseline~\cite{TATS}.
We set the size of the latent bottleneck to $N_L=256$ for every experiment.
We train 16-frame MeBT without curriculum while 128-frame MeBT models are trained with the proposed learning schedule in Section~\ref{sec:train_inference}. 
For the interval scheduling function, we use $\beta = 2$ for all datasets while $\alpha = 30K$ for SkyTimelapse and $\alpha = 100K$ for Taichi-HD and UCF-101. 
More details on training are described in Appendix (Section \ref{appx:training_inference}).

\subsection{Results on Short-term Video Generation}
\paragraph{Baselines}
We compare our method with state-of-the-art video generative models whose pre-trained models are publicly available.
In all datasets, we employ MoCoGAN-HD~\cite{MoCoGAN-HD} and DIGAN~\cite{DIGAN} as methods based on Generative Adversarial Networks (GAN), and TATS~\cite{TATS} as the state-of-the-art autoregressive transformer that differs only in architecture while sharing the same quantization codebook with ours.
We also present comparisons to other unconditional models in UCF-101 using the reported numbers.

\cutparagraphup
\paragraph{Results}
Table~\ref{tab:short-quantitative} summarizes the results on short-term video generation (16 frames).
Overall, we observe that our method exhibits competitive performance to the state-of-the-art approaches.
Our method outperforms all non-transformer baselines by a large margin in all datasets, and competitive performance to other autoregressive transformers (TATS and CCVS); it achieves the best and the second-best performance in SkyTimelapse and Taichi-HD, respectively, and the best Inception score in UCF-101.
Compared to TATS, however, our method exhibits considerable improvement in efficiency in both training and inference; it reduces the peak training memory usage by $33\%$ while improving the inference speed by $10$ times.
Note that the gap increases rapidly in longer videos as the asymptotic complexity is quadratic in TATS but linear in MeBT.
Overall, our results show that MeBT provides compelling short-term generation performance while reducing computation cost significantly. 
It allows our model to be trained with longer videos to model interesting long-term dynamics, which is described in the next section.

\begin{figure}[ht]
    \centering
    \includegraphics[width=0.47\textwidth]{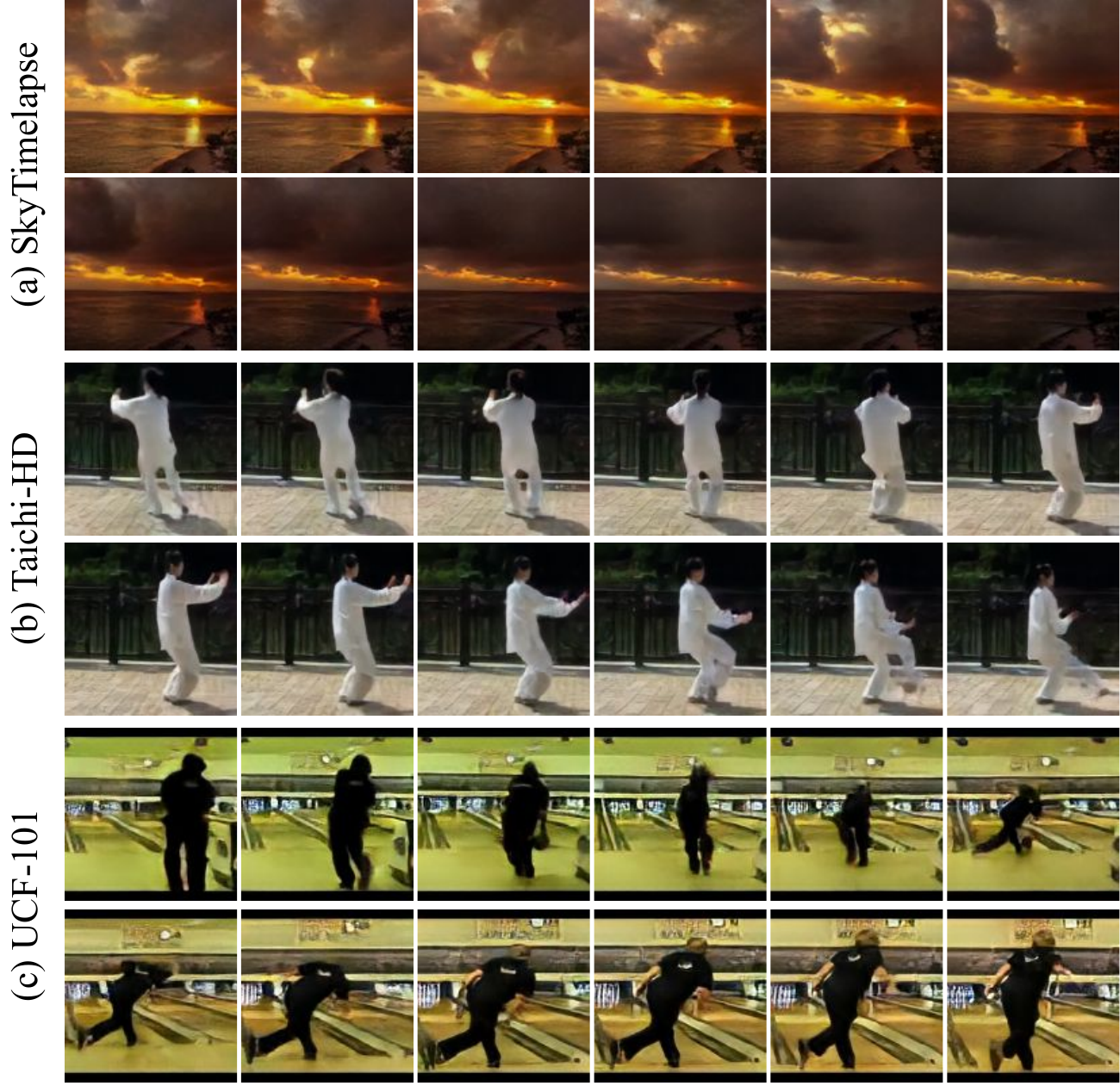}
    \vspace{-0.17cm}
    \caption{
    Qualitative results of MeBT on 128-frame videos. We present every 16th frames from the generated videos.
    }
    \label{fig:longterm_ours}
    \vspace{-0.7cm}
\end{figure}

\begin{figure*}[ht]
    \centering
    \includegraphics[width=0.9\textwidth]{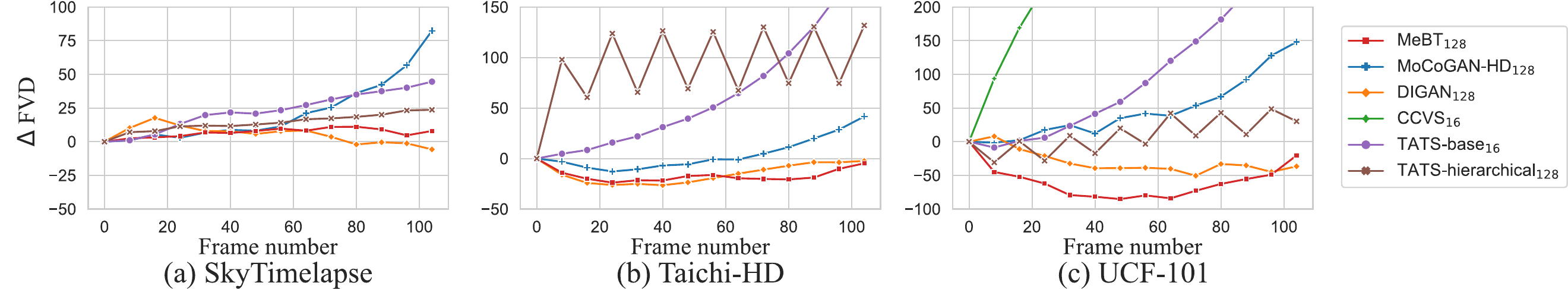}
    \vspace{-0.3cm}
    \caption{Evaluation of generation quality over time. FVD is measured in 16-frame interval relative to the initial prediction.}
    \label{fig:error_propagation}
    \vspace{-0.5cm}
\end{figure*}

\subsection{Results on Long-term Video Generation}
\label{sec:exp_long_video}
\paragraph{Baselines}
We evaluate our method with five strong baselines: MoCoGAN-HD~\cite{MoCoGAN-HD}, CCVS~\cite{CCVS}\footnote{For CCVS, we report the results only in UCF-101 dataset where the official pre-trained model is available.}, and TATS~\cite{TATS} as strong autoregressive approaches that recursively generate long videos conditioned on the previous generation, and DIGAN~\cite{DIGAN} as an implicit network that directly generates each frame independently conditioned on latent variable. 
We also compare with the hierarchical variants of TATS~\cite{TATS} that generates videos in coarse-to-fine manner using two separate transformers; one generates keyframes in coarse temporal scale, and the other interpolates missing frames between two consecutive keyframes.
We trained the baselines on 128-frame videos if their training peak memory stays under the VRAM capacity, and used 16-frames otherwise.
See the Appendix for more details on the baselines (Section \ref{appx:baselines}).

\cutparagraphup
\paragraph{Results}
Table~\ref{tab:longterm} and Figure~\ref{fig:longterm_qualitative} summarize the quantitative and qualitative comparisons on long-term video synthesis, respectively.
The quantitative results show that our method outperforms all baselines by a large margin over all datasets.
More specifically, the qualitative results display that the prediction error accumulates over time in na\"ive autoregressive transformers (Fig.~\ref{fig:longterm_qualitative}(b)), which is partly addressed in the hierarchical method yet suffering from the inconsistency at keyframe boundaries and inaccurate interpolation (Fig.~\ref{fig:longterm_qualitative}(c)). 
The DIGAN based on implicit representation exhibits more robustness to error propagation due to non-autoregressive generation process, yet fails to capture meaningful structures in complex videos (Fig.~\ref{fig:longterm_qualitative}(d)).
Compared to these methods, MeBT generates convincing videos by generating not only high-fidelity frames consistently over time but also coherent semantic structures and dynamics, such as a person leaving and re-entering the scene (Fig.~\ref{fig:longterm_qualitative}(a)), by exploiting long-term dependencies.
Figure~\ref{fig:longterm_ours} illustrates more qualitative results of our method.
It shows that MeBT can capture interesting long-term dynamics, such as global dynamics as the sun goes down over the water (Fig.~\ref{fig:longterm_ours}(a)), complex dynamics of a human (Fig.~\ref{fig:longterm_ours}(b)) and combination of human and camera motion (Fig.~\ref{fig:longterm_ours}(c)).

To assess the consistency of generation quality, we measure FVD over time relative to the initial predictions and summarize the results in Figure~\ref{fig:error_propagation}.
The autoregressive models (MoCoGAN-HD, CCVS, and TATS-base) suffer from dramatic error propagation. Conversely, the hierarchical model (TATS-hierarchical) is plagued by inconsistencies around keyframe boundaries that manifest as zigzag patterns. In contrast, DIGAN and MeBT achieve near-zero quality degradation in all datasets ($\vartriangle \text{FVD}\approx 0$) due to their non-autoregressive design and end-to-end training with long sequences. Nevertheless, unlike MeBT, DIGAN fails to generate meaningful structures as discussed in previous paragraphs (Table~\ref{tab:longterm}).

\subsection{Ablation Studies}
\label{sec:ablation_study}
This section presents component-wise analysis of MeBT. 
All experiments are conducted with SkyTimelapse videos. Except for the ablation studies about long-range dependency modeling, we used 16-frame model for simplicity.

\cutparagraphup
\paragraph{Impact of latent bottleneck size}
To study the efficiency and performance trade-off introduced by the latent bottleneck, we train MeBT with varying number of latent codes and report the results in Table~\ref{tab:16f_ablation}.
Since the entire context tokens are compressed into much smaller latents by the encoder (Section~\ref{sec:mebt_encoder}), employing small latents generally sacrifices the performance. 
Yet, the performance quickly saturates at a reasonable scale $N_L=256$, which is still considerably smaller than the context size $N_C=1024$.
It shows that MeBT can achieve linear complexity without compromising much performance.

\begin{table}[t]
    \caption{Ablation study on decoder design and bottleneck sizes.}
    \vspace{-0.3cm}
    \label{tab:16f_ablation}
    \centering
    \footnotesize
    \begin{tabular}{cc|ccc}
    \toprule
    Decoder & $N_L$& Memory & FVD$_{16}$ & KVD$_{16}$ \\
    \midrule
    Full & 128 & 18.2 GB & { 92.2\tiny{$\pm2.2$}} & {\bf 2.6\tiny{$\pm.12$}} \\
    \hline
    Ours & 32 & {\bf 6.0} GB & 110.8\tiny{$\pm2.4$} & 4.0\tiny{$\pm.15$} \\
    Ours & 128 & {7.4} GB & 95.2\tiny{$\pm1.5$} & 2.8\tiny{$\pm.11$} \\
    Ours & 256 & {9.5} GB & {\bf 90.4}\tiny{$\pm1.8$} & 2.8\tiny{$\pm.17$} \\
    Ours & 512 & {14.7} GB & 90.8\tiny{$\pm2.1$} & \bf{2.6}\tiny{$\pm.14$} \\
    \bottomrule
    \end{tabular}
    \vspace{-0.3cm}
\end{table}

\cutparagraphup
\paragraph{Impact of the decoder architecture}
\label{sec:dec_ablation}
To evaluate the impact of our decoder (Section~\ref{sec:mebt_decoder}), we compare our method with the bidirectional decoder that leverages the full self-attention among the latent and masked tokens.
Comparing the first and third rows in Table~\ref{tab:16f_ablation} shows that approximating all pair-wise interaction of context and mask tokens through linear-time MeBT decoder merely degrades the performance while considerably reducing the peak memory.

\begin{table}[t]
\caption{Ablation study on attention and scheduling functions. The training peak memory is measured with batch size 1.}
\label{tab:128f_ablation}
\vspace{-0.3cm}
    \centering
    \footnotesize
    \begin{tabular}{cc|cccc}
    \toprule
        Model & Schedule & FVD$_{128}$ & KVD$_{128}$ & Memory & Time\\
    \midrule
        Window & Gaussian & 458\tiny$\pm21.2$ & 33.2\tiny$\pm3.31$ & 34.9 GB & 13.7s\\
        Axial & Gaussian & 377\tiny$\pm12.2$& 13.6\tiny$\pm0.81$ & 21.2 GB & 12.0s \\
        \hline
        Ours & None & 276\tiny{$\pm10.9$} & 7.5\tiny{$\pm.99$} & \textbf{13.3} GB & \textbf{6.53}s\\
        Ours & Uniform & 253\tiny{$\pm4.6$} & 6.7\tiny{$\pm.49$} & \textbf{13.3} GB & \textbf{6.53}s\\
        \hline
        Ours & Gaussian & \textbf{239\tiny${\pm3.6}$} & \textbf{5.1}\tiny$\pm0.16$ & \textbf{13.3} GB & \textbf{6.53}s\\
    \bottomrule
    \end{tabular}
\vspace{-0.5cm}
\end{table}

\cutparagraphup
\paragraph{Impact of attention structure}
We evaluate the impact of attention structure on 128-frame videos, by replacing the latent bottleneck attention in MeBT with sparse attention methods used in video modeling: axial attention~\cite{Ho2019} alternates attention over horizontal, vertical, and temporal axes, and window attention~\cite{MaskViT} alternates attention over spatial and temporal axes.
See Section~\ref{appx:asymptotic_mem} for more details on the baselines.
As shown in Table \ref{tab:128f_ablation}, ours shows clear improvements over the baselines in both efficiency and performance.
In terms of efficiency, MeBT outperforms baselines in both memory and inference time thanks to its linear complexity, while baselines have quadratic complexity to the temporal length from self-attention on the temporal axis.
Interestingly, MeBT also exhibits improvement in sample quality since the heuristic sparsity patterns in the baselines are insufficient to capture complex long-range dependency while ours can \emph{learn} them adaptively through the latents.

\cutparagraphup
\paragraph{Impact of interval scheduling function}
To evaluate the effectiveness of the interval scheduling function in Section~\ref{sec:train_inference}, we train MeBT on 128-frame videos with three different stratagies: \emph{None} refers random sampling over entire video volume ($p_n(v)=\delta(n=t)$), \emph{Uniform} refers sampling the video interval uniformly random ($p_n(v)=\frac{1}{t}$), and the \emph{Gaussian} refers the proposed curriculum sampling in Eq.~\eqref{eq:masking_schedule}.
As summarized in Table~\ref{tab:128f_ablation}, sampling the context and mask tokens randomly from long videos often leads to suboptimal results (\emph{None}), while limiting the interval of the samples makes the training easier (\emph{Uniform}). 
By learning from short to long videos (\emph{Gaussian}), the proposed strategy stabilizes the training and improves the performance.

\cutparagraphup
\section{Conclusion}
We proposed a Memory-efficient Bidirectional Transformer (MeBT), a transformer-based generative model for moderately long-term video synthesis.
By formulating the video synthesis as an iterative mask prediction task and employing bidirectional transformers with efficient encoder-decoder architecture, we showed that we could push the transformers to leverage much longer videos in training while enjoying fast inference speed and robustness in error propagation.
Our experiments demonstrated that by training with longer videos, MeBT could learn long-term dependencies and generate coherent videos over longer time horizon.

\cutparagraphup
\paragraph{Acknowledgements}
This work was supported in part by the Institute of Information \& communications Technology Planning \& Evaluation (IITP) (No.  and 2019-0-00075) and the National Research Foundation of Korea (NRF) (No. 2021R1C1C1012540 and 2021R1A4A3032834) funded by the Korea government (MSIT), Korea Meteorological Administration under Grant (KMA2021-00121), and Kakao Brain corporations. 
We thank Jinwoo Kim (KAIST) for helpful comments and discussions.

{\small
\bibliographystyle{ieee_fullname}
\bibliography{egbib}

\begin{thebibliography}{10}\itemsep=-1pt

\bibitem{Akan2022}
Adil~Kaan Akan, Sadra Safadoust, Erkut Erdem, Aykut Erdem, and Fatma
  G{\"{u}}ney.
\newblock Stochastic video prediction with structure and motion.
\newblock {\em CoRR}, abs/2203.10528, 2022.

\bibitem{FitVid}
Mohammad Babaeizadeh, Mohammad~Taghi Saffar, Suraj Nair, Sergey Levine, Chelsea
  Finn, and Dumitru Erhan.
\newblock Fitvid: Overfitting in pixel-level video prediction.
\newblock {\em CoRR}, abs/2106.13195, 2021.

\bibitem{LongFormer}
Iz Beltagy, Matthew~E. Peters, and Arman Cohan.
\newblock Longformer: The long-document transformer.
\newblock {\em CoRR}, abs/2004.05150, 2020.

\bibitem{ImprovedVRNN}
Llu{\'{\i}}s Castrej{\'{o}}n, Nicolas Ballas, and Aaron~C. Courville.
\newblock Improved conditional vrnns for video prediction.
\newblock In {\em ICCV}, 2019.

\bibitem{MaskGIT}
Huiwen Chang, Han Zhang, Lu Jiang, Ce Liu, and William~T. Freeman.
\newblock Maskgit: Masked generative image transformer.
\newblock In {\em CVPR}, 2022.

\bibitem{DVDGAN}
Aidan Clark, Jeff Donahue, and Karen Simonyan.
\newblock Efficient video generation on complex datasets.
\newblock {\em CoRR}, abs/1907.06571, 2019.

\bibitem{SVG}
Emily Denton and Rob Fergus.
\newblock Stochastic video generation with a learned prior.
\newblock In {\em ICML}, 2018.

\bibitem{BERT}
Jacob Devlin, Ming{-}Wei Chang, Kenton Lee, and Kristina Toutanova.
\newblock {BERT:} pre-training of deep bidirectional transformers for language
  understanding.
\newblock In {\em NAACL-HLT}, 2019.

\bibitem{VQGAN}
Patrick Esser, Robin Rombach, and Bj{\"{o}}rn Ommer.
\newblock Taming transformers for high-resolution image synthesis.
\newblock In {\em CVPR}, 2021.

\bibitem{SRVP}
Jean{-}Yves Franceschi, Edouard Delasalles, Micka{\"{e}}l Chen, Sylvain
  Lamprier, and Patrick Gallinari.
\newblock Stochastic latent residual video prediction.
\newblock In {\em ICML}, 2020.

\bibitem{TATS}
Songwei Ge, Thomas Hayes, Harry Yang, Xi Yin, Guan Pang, David Jacobs,
  Jia{-}Bin Huang, and Devi Parikh.
\newblock Long video generation with time-agnostic {VQGAN} and time-sensitive
  transformer.
\newblock In {\em ECCV}, 2022.

\bibitem{MaskViT}
Agrim Gupta, Stephen Tian, Yunzhi Zhang, Jiajun Wu, Roberto
  Mart{\'{\i}}n{-}Mart{\'{\i}}n, and Li Fei{-}Fei.
\newblock Maskvit: Masked visual pre-training for video prediction.
\newblock {\em CoRR}, abs/2206.11894, 2022.

\bibitem{MMVID}
Ligong Han, Jian Ren, Hsin{-}Ying Lee, Francesco Barbieri, Kyle Olszewski,
  Shervin Minaee, Dimitris~N. Metaxas, and Sergey Tulyakov.
\newblock Show me what and tell me how: Video synthesis via multimodal
  conditioning.
\newblock In {\em CVPR}, 2022.

\bibitem{MAE}
Kaiming He, Xinlei Chen, Saining Xie, Yanghao Li, Piotr Doll{\'{a}}r, and
  Ross~B. Girshick.
\newblock Masked autoencoders are scalable vision learners.
\newblock In {\em CVPR}, 2022.

\bibitem{Ho2019}
Jonathan Ho, Nal Kalchbrenner, Dirk Weissenborn, and Tim Salimans.
\newblock Axial attention in multidimensional transformers.
\newblock {\em CoRR}, abs/1912.12180, 2019.

\bibitem{CogVideo}
Wenyi Hong, Ming Ding, Wendi Zheng, Xinghan Liu, and Jie Tang.
\newblock Cogvideo: Large-scale pretraining for text-to-video generation via
  transformers.
\newblock {\em CoRR}, abs/2205.15868, 2022.

\bibitem{PerceiverIO}
Andrew Jaegle, Sebastian Borgeaud, Jean{-}Baptiste Alayrac, Carl Doersch,
  Catalin Ionescu, David Ding, Skanda Koppula, Daniel Zoran, Andrew Brock, Evan
  Shelhamer, Olivier~J. H{\'{e}}naff, Matthew~M. Botvinick, Andrew Zisserman,
  Oriol Vinyals, and Jo{\~{a}}o Carreira.
\newblock Perceiver {IO:} {A} general architecture for structured inputs {\&}
  outputs.
\newblock In {\em ICLR}, 2022.

\bibitem{Perceiver}
Andrew Jaegle, Felix Gimeno, Andy Brock, Oriol Vinyals, Andrew Zisserman, and
  Jo{\~{a}}o Carreira.
\newblock Perceiver: General perception with iterative attention.
\newblock In {\em ICML}, 2021.

\bibitem{LDVDGAN}
Emmanuel Kahembwe and Subramanian Ramamoorthy.
\newblock Lower dimensional kernels for video discriminators.
\newblock {\em Neural Networks}, 2020.

\bibitem{kinetics}
Will Kay, Jo{\~{a}}o Carreira, Karen Simonyan, Brian Zhang, Chloe Hillier,
  Sudheendra Vijayanarasimhan, Fabio Viola, Tim Green, Trevor Back, Paul
  Natsev, Mustafa Suleyman, and Andrew Zisserman.
\newblock The kinetics human action video dataset.
\newblock {\em CoRR}, abs/1705.06950, 2017.

\bibitem{SetVAE}
Jinwoo Kim, Jaehoon Yoo, Juho Lee, and Seunghoon Hong.
\newblock Setvae: Learning hierarchical composition for generative modeling of
  set-structured data.
\newblock In {\em CVPR}, 2021.

\bibitem{KimNCK19}
Yunji Kim, Seonghyeon Nam, In Cho, and Seon~Joo Kim.
\newblock Unsupervised keypoint learning for guiding class-conditional video
  prediction.
\newblock In {\em NeurIPS}, 2019.

\bibitem{SAVP}
Alex~X. Lee, Richard Zhang, Frederik Ebert, Pieter Abbeel, Chelsea Finn, and
  Sergey Levine.
\newblock Stochastic adversarial video prediction.
\newblock {\em CoRR}, abs/1804.01523, 2018.

\bibitem{draft_and_revise}
Doyup Lee, Chiheon Kim, Saehoon Kim, Minsu Cho, and Wook-Shin Han.
\newblock Draft-and-revise: Effective image generation with contextual
  rq-transformer.
\newblock In {\em NeurIPS}, 2022.

\bibitem{SetTransformer}
Juho Lee, Yoonho Lee, Jungtaek Kim, Adam~R. Kosiorek, Seungjin Choi, and
  Yee~Whye Teh.
\newblock Set transformer: {A} framework for attention-based
  permutation-invariant neural networks.
\newblock In Kamalika Chaudhuri and Ruslan Salakhutdinov, editors, {\em ICML},
  2019.

\bibitem{LeeKCKR21}
Sangmin Lee, Hak~Gu Kim, Dae~Hwi Choi, Hyung{-}Il Kim, and Yong~Man Ro.
\newblock Video prediction recalling long-term motion context via memory
  alignment learning.
\newblock In {\em CVPR}, 2021.

\bibitem{HVP}
Wonkwang Lee, Whie Jung, Han Zhang, Ting Chen, Jing~Yu Koh, Thomas~E. Huang,
  Hyungsuk Yoon, Honglak Lee, and Seunghoon Hong.
\newblock Revisiting hierarchical approach for persistent long-term video
  prediction.
\newblock In {\em ICLR}, 2021.

\bibitem{LiangLDX17}
Xiaodan Liang, Lisa Lee, Wei Dai, and Eric~P. Xing.
\newblock Dual motion {GAN} for future-flow embedded video prediction.
\newblock In {\em ICCV}, 2017.

\bibitem{AdamW}
Ilya Loshchilov and Frank Hutter.
\newblock Decoupled weight decay regularization.
\newblock In {\em ICLR}, 2019.

\bibitem{LUNA}
Xuezhe Ma, Xiang Kong, Sinong Wang, Chunting Zhou, Jonathan May, Hao Ma, and
  Luke Zettlemoyer.
\newblock Luna: Linear unified nested attention.
\newblock In {\em NeurIPS}, 2021.

\bibitem{CCVS}
Guillaume~Le Moing, Jean Ponce, and Cordelia Schmid.
\newblock {CCVS:} context-aware controllable video synthesis.
\newblock In {\em NeurIPS}, 2021.

\bibitem{LVT}
Ruslan Rakhimov, Denis Volkhonskiy, Alexey Artemov, Denis Zorin, and Evgeny
  Burnaev.
\newblock Latent video transformer.
\newblock In {\em VISIGRAPP}, 2021.

\bibitem{TGAN}
Masaki Saito, Eiichi Matsumoto, and Shunta Saito.
\newblock Temporal generative adversarial nets with singular value clipping.
\newblock In {\em ICCV}, 2017.

\bibitem{TGANv2}
Masaki Saito, Shunta Saito, Masanori Koyama, and Sosuke Kobayashi.
\newblock Train sparsely, generate densely: Memory-efficient unsupervised
  training of high-resolution temporal {GAN}.
\newblock {\em Int. J. Comput. Vis.}, 2020.

\bibitem{IS}
Tim Salimans, Ian~J. Goodfellow, Wojciech Zaremba, Vicki Cheung, Alec Radford,
  and Xi Chen.
\newblock Improved techniques for training gans.
\newblock In {\em NeurIPS}, 2016.

\bibitem{CWVAE}
Vaibhav Saxena, Jimmy Ba, and Danijar Hafner.
\newblock Clockwork variational autoencoders.
\newblock In {\em NeurIPS}, 2021.

\bibitem{taichi}
Aliaksandr Siarohin, St{\'{e}}phane Lathuili{\`{e}}re, Sergey Tulyakov, Elisa
  Ricci, and Nicu Sebe.
\newblock First order motion model for image animation.
\newblock In {\em NeurIPS}, 2019.

\bibitem{StyleGAN-V}
Ivan Skorokhodov, Sergey Tulyakov, and Mohamed Elhoseiny.
\newblock Stylegan-v: {A} continuous video generator with the price, image
  quality and perks of stylegan2.
\newblock In {\em CVPR}, 2022.

\bibitem{ucf}
Khurram Soomro, Amir~Roshan Zamir, and Mubarak Shah.
\newblock {UCF101:} {A} dataset of 101 human actions classes from videos in the
  wild.
\newblock {\em CoRR}, abs/1212.0402, 2012.

\bibitem{EfficientTransformers}
Yi Tay, Mostafa Dehghani, Dara Bahri, and Donald Metzler.
\newblock Efficient transformers: {A} survey.
\newblock {\em arXiv}, 2020.

\bibitem{MoCoGAN-HD}
Yu Tian, Jian Ren, Menglei Chai, Kyle Olszewski, Xi Peng, Dimitris~N. Metaxas,
  and Sergey Tulyakov.
\newblock A good image generator is what you need for high-resolution video
  synthesis.
\newblock In {\em ICLR}, 2021.

\bibitem{c3d}
Du Tran, Lubomir~D. Bourdev, Rob Fergus, Lorenzo Torresani, and Manohar Paluri.
\newblock Learning spatiotemporal features with 3d convolutional networks.
\newblock In {\em ICCV}, 2015.

\bibitem{MoCoGAN}
Sergey Tulyakov, Ming{-}Yu Liu, Xiaodong Yang, and Jan Kautz.
\newblock Mocogan: Decomposing motion and content for video generation.
\newblock In {\em CVPR}, 2018.

\bibitem{FVD}
Thomas Unterthiner, Sjoerd van Steenkiste, Karol Kurach, Rapha{\"{e}}l
  Marinier, Marcin Michalski, and Sylvain Gelly.
\newblock Towards accurate generative models of video: {A} new metric {\&}
  challenges.
\newblock {\em CoRR}, abs/1812.01717, 2018.

\bibitem{VQVAE}
A{\"{a}}ron van~den Oord, Oriol Vinyals, and Koray Kavukcuoglu.
\newblock Neural discrete representation learning.
\newblock In {\em NeurIPS}, 2017.

\bibitem{Villegas2017}
Ruben Villegas, Jimei Yang, Yuliang Zou, Sungryull Sohn, Xunyu Lin, and Honglak
  Lee.
\newblock Learning to generate long-term future via hierarchical prediction.
\newblock In Doina Precup and Yee~Whye Teh, editors, {\em ICML}, 2017.

\bibitem{VGAN}
Carl Vondrick, Hamed Pirsiavash, and Antonio Torralba.
\newblock Generating videos with scene dynamics.
\newblock In {\em NeurIPS}, 2016.

\bibitem{VPVQVAE}
Jacob Walker, Ali Razavi, and A{\"{a}}ron van~den Oord.
\newblock Predicting video with {VQVAE}.
\newblock {\em CoRR}, abs/2103.01950, 2021.

\bibitem{Linformer}
Sinong Wang, Belinda~Z. Li, Madian Khabsa, Han Fang, and Hao Ma.
\newblock Linformer: Self-attention with linear complexity.
\newblock {\em arXiv}, 2020.

\bibitem{GHVAE}
Bohan Wu, Suraj Nair, Roberto Mart{\'{\i}}n{-}Mart{\'{\i}}n, Li Fei{-}Fei, and
  Chelsea Finn.
\newblock Greedy hierarchical variational autoencoders for large-scale video
  prediction.
\newblock In {\em CVPR}, 2021.

\bibitem{NUWA}
Chenfei Wu, Jian Liang, Lei Ji, Fan Yang, Yuejian Fang, Daxin Jiang, and Nan
  Duan.
\newblock N{\"{u}}wa: Visual synthesis pre-training for neural visual world
  creation.
\newblock In {\em ECCV}, 2022.

\bibitem{stl}
Wei Xiong, Wenhan Luo, Lin Ma, Wei Liu, and Jiebo Luo.
\newblock Learning to generate time-lapse videos using multi-stage dynamic
  generative adversarial networks.
\newblock In {\em CVPR}, 2018.

\bibitem{videoGPT}
Wilson Yan, Yunzhi Zhang, Pieter Abbeel, and Aravind Srinivas.
\newblock Videogpt: Video generation using vq-vae and transformers.
\newblock {\em arXiv preprint arXiv:2104.10157}, 2021.

\bibitem{DIGAN}
Sihyun Yu, Jihoon Tack, Sangwoo Mo, Hyunsu Kim, Junho Kim, Jung{-}Woo Ha, and
  Jinwoo Shin.
\newblock Generating videos with dynamics-aware implicit generative adversarial
  networks.
\newblock In {\em ICLR}, 2022.

\bibitem{BigBird}
Manzil Zaheer, Guru Guruganesh, Kumar~Avinava Dubey, Joshua Ainslie, Chris
  Alberti, Santiago Onta{\~{n}}{\'{o}}n, Philip Pham, Anirudh Ravula, Qifan
  Wang, Li Yang, and Amr Ahmed.
\newblock Big bird: Transformers for longer sequences.
\newblock In {\em NeurIPS}, 2020.

\bibitem{M6-UFC}
Zhu Zhang, Jianxin Ma, Chang Zhou, Rui Men, Zhikang Li, Ming Ding, Jie Tang,
  Jingren Zhou, and Hongxia Yang.
\newblock {UFC-BERT:} unifying multi-modal controls for conditional image
  synthesis.
\newblock In {\em NeurIPS}, 2021.

\end{thebibliography}
}
\clearpage
\appendix

\section*{Appendix}
This document provides comprehensive descriptions and results of our method that could not be accommodated in the main paper due to space restriction. More generated samples from MeBT and the baselines~\cite{MoCoGAN-HD, DIGAN, TATS, CCVS} can be found at \href{https://sites.google.com/view/mebt-cvpr2023}{https://sites.google.com/view/mebt-cvpr2023}.

\section{Experimental Details}
\label{appx:experiment}
In this section, we first describe the detailed implementation of training and inference in Section~\ref{sec:train_inference} in the main paper.
Then, we provide additional descriptions of the datasets and baselines used in Section~\ref{sec:experiments} in the main paper.

\subsection{Training and Inference}
\label{appx:training_inference}
\paragraph{Training}
\label{appx:training}
For all experiments, we used the same network architecture except for the size of latent bottleneck $N_L$. Following the configuration of TATS~\cite{TATS}, all models have 24 attention layers, 16 attention heads, 1024 embedding dimensions, and learnable positional embedding for all spatiotemporal positions. For the size of latent bottleneck, we used $N_L=256$ for all experiments.
We use AdamW optimizer~\cite{AdamW} with $\beta_1 = 0.9$ and $\beta_2 = 0.95$. Table~\ref{tab:train_config} summarizes our training configuration for all experiments.

\paragraph{Inference}
\label{appx:inference}
When sampling long videos, we decode $N_d$ tokens by appending a single token to the context tokens $N_d$ times before applying the iterative decoding to hold the consistency of the video. Then, the iterative decoding is applied with the $N_d$ decoded context tokens. For the iterative decoding, we adopt a top-k sampling strategy when sampling the code from logits. We also adopt context temperature annealing when updating the context tokens to increase the diversity of samples following the official implementation of MaskGIT\footnote{https://github.com/google-research/maskgit}. Specifically, let $p_i \in (0, 1)$ be the probability value of sampled token $i$, and $s$, $S$ be the current decoding step and total decoding steps, respectively. Then, we update context tokens with top-$N_s$ decoded tokens with the highest confidence scores $c_i~=~p_i +~\text{Gumbel(0,1)} * (1-\frac{s}{S})\tau$. For the revision, we repeated the revision $N_R$ times with the logits scaled with a certain temperature. The specific configuration for each experiment is summarized in Table \ref{tab:inference_config}.

\subsection{Datasets}
\label{appx:datasets}
\paragraph{SkyTimelapse} SkyTimelapse~\cite{stl} contains time-lapse videos that illustrate the dynamics of the sky such as the growth and motions of clouds. We used the training split to train the model and the validation split to measure the FVD. We train our model with videos that are longer than the training length. We utilized 2,115 videos for training the short-term models and 1,059 videos for training the long-term models.

\paragraph{Taichi-HD} Taichi-HD is a collection of videos of a single actor performing Taichi. Taichi-HD has a total of 2,854 videos including both training and validation split. For Taichi-HD, we utilize both splits and sampled every 4 frames when training on 16-frame videos following the setting of \cite{TATS, DIGAN}. For 128-frame videos, we sampled all frames for training as the 4-frame sampling drops 2,438 videos over 2,854 videos. By sampling every frame, we could utilize 2,760 videos for training the 128-frame model.

\paragraph{UCF-101} UCF-101 is an action recognition dataset that contains 13,320 videos with 101 classes in total. We trained our short-term model on the training split with 9,537 16-frame videos and our long-term model with 6,469 128-frame videos.

\subsection{Long-term Video Generation Baselines}
\label{appx:baselines}
\paragraph{MoCoGAN-HD} MoCoGAN-HD~\cite{MoCoGAN-HD} models videos by training a motion generator on the image latent space provided by a pre-trained image generator. The motion generator of MoCoGAN-HD is implemented with RNNs and thus has sub-quadratic complexity to the video length. Therefore, we trained MoCoGAN-HD with 128-frame videos for long-term comparison.

\paragraph{DIGAN}
DIGAN~\cite{DIGAN} considers a video as a function that outputs the RGB pixel value for the given spatiotemporal coordinate. DIGAN models videos with a motion discriminator that achieves constant complexity by discovering unnatural motions when two frames and the time gap between frames are given. As the discriminator gets two frames and the video is generated by a coordinate-based network, DIGAN has sub-quadratic complexity to the video length. Thus, we trained DIGAN with 128-frame videos for long-term comparison.

\paragraph{CCVS}
CCVS~\cite{CCVS} is a video prediction model that utilizes optical flow and an autoregressive transformer. CCVS can generate longer videos by sliding the attention window, but due to the quadratic complexity of the autoregressive transformer, it cannot be directly trained with long videos.
Hence, for the long-term comparison, we utilized the 16-frame CCVS model to predict 128-frame videos from randomly sampled real frames as an initial frame.

\paragraph{TATS}
TATS~\cite{TATS} models videos by adopting an autoregressive transformer and a time-agnostic 3d VQGAN. The proposed time-agnostic 3d VQGAN can decode longer videos beyond the training length. Therefore, TATS-base can generate longer videos by sliding the attention window. However, due to the quadratic complexity of transformers, TATS-base cannot directly model long videos. To address this issue, the authors proposed TATS-hierarchical that generates long videos sparsely and interpolates the missing frames. We compared TATS-base by applying a sliding attention window, and TATS-hierarchical by training the hierarchical model on 128-frame videos.

\subsection{Comparison to Sparse Attentions}
\label{appx:asymptotic_mem}

This section presents more in-depth analysis and comparisons with the baselines presented in Section~\ref{sec:ablation_study}.
Let $N(=H\times W)$ and $T$ denote the number of tokens in spatial and temporal dimensions, respectively. Then,
\begin{enumerate}
    \item {\bf Axial attention}~\cite{Ho2019} alternates attention over horizontal, vertical, and temporal axes. 
    The queries in each attention layer can only interact with tokens on the same axis. 
    The overall complexity is $O(NT(H+W+T))$.
    \item {\bf Window attention}~\cite{MaskViT} alternates attention over spatial and temporal axes. It performs frame-wise attention over $N$ tokens followed by spatio-temporal attention over $n\times T$ tokens where $n(=16)$ is the size of spatial window. The complexity is $O(N^2T+nNT^2)$.
    \vspace{-0.4cm}
    \item {\bf Local attention}~\cite{NUWA} computes the attention within a fixed number ($n$) of spatio-temporal neighborhood tokens. The complexity of local attention is $O(nNT)$.
\end{enumerate}

While axial attention and window attention are asymptotically more efficient than dense attention of $O(N^2T^2)$, they still have quadratic complexity with respect to the spatial ($N$) or temporal dimension ($T$). On the other hand, local attention and MeBT with $n$ latent codes have a linear complexity of $O(nNT)$ for both N and T. 
However, local attention cannot effectively model the long-term dependency of tokens due to the limited receptive fields, and its benefits are often not fully leveraged due to the lacking support in both hardware and software for the sparse operations\footnote{Most of the currently available implementations of local attention is not leveraging the sparse operators, which makes them have the same complexity as dense attention.}. 
Therefore, we only compared MeBT with axial and window attention in our ablation study in Section~\ref{sec:ablation_study}.

\begin{figure}[t]
    \centering
    \includegraphics[width=0.4\textwidth]{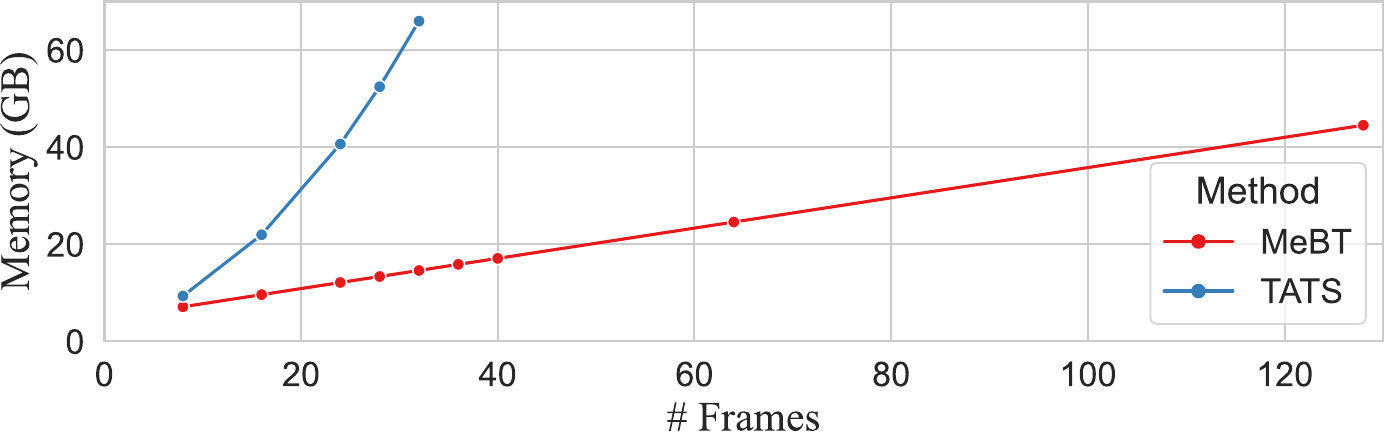}
    \caption{Empirically measured training peak memory over training video lengths. MeBT showed linear complexity to the video length, while TATS showed quadratic complexity.}
    \label{fig:memory_comparison}
\end{figure}
\section{Empirical Analysis on Memory Complexity}
\label{appx:empirical_mem}
We compare MeBT's empirical memory consumption over video lengths with an autoregressive transformer by measuring the training peak memory on a single A100 GPU with 80GB of VRAM with batch size 4. The results are shown in Figure \ref{fig:memory_comparison}. As shown in the figure, the training peak memory of MeBT scaled up linearly while the autoregressive transformer \cite{TATS} showed quadratic growth. In particular, with batch size 4, we couldn't train the autoregressive transformer with videos longer than 36 frames represented with 2304 tokens due to the out-of-memory error. Compared to the autoregressive transformer, our model spends about 40GB when training with 128-frame videos (8,192 tokens). It shows that MeBT is much more efficient than the transformer in long videos and capable of modeling longer videos directly.

\section{Additional Qualitative Results}
\label{appx:qualitative}
In this section, we extend the discussion in Section~\ref{sec:exp_long_video} in the main paper with additional qualitative results.
The generated videos from MeBT and the baselines can be found at \href{https://sites.google.com/view/mebt-cvpr2023}{https://sites.google.com/view/mebt-cvpr2023}.
\subsection{Qualitative Comparison on SkyTimelapse}
\begin{figure*}[t]
    \centering
    \includegraphics[width=1.0\textwidth]{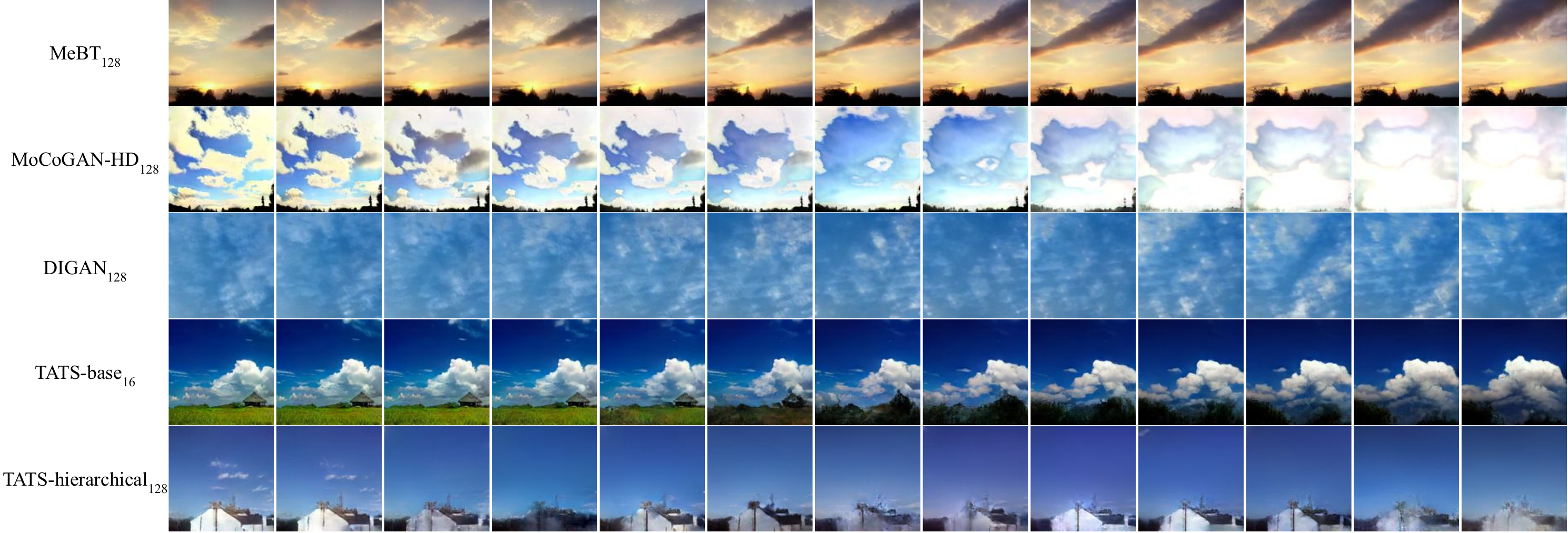}
    \caption{Generated videos on SkyTimelapse. We displayed every 10th frame in the generated video. The subscript denotes the length of training videos. More samples can be found on the website.}
    \label{fig:supp_stl}
\end{figure*}

The generated videos from the baselines and MeBT are displayed in Figure \ref{fig:supp_stl}. As shown in the figure, our model demonstrates consistent high-fidelity sky videos. To be specific, compared to TATS-base and TATS-hierarchical, our model showed better consistency on modeling the static ground. This is because our model can decode the tokens in a bidirectional manner. As autoregressive transformers follow the raster-scan ordering, the tokens that represent the ground in the previous frame and current frame are far away on the sequence. Compared to DIGAN~\cite{DIGAN} and MoCoGAN-HD~\cite{MoCoGAN-HD}, MeBT exhibits high-fidelity video generation while DIGAN shows unrealistic artifacts and MoCoGAN-HD collapses when generating long-term frames.

\subsection{Qualitative Comparison on Taichi-HD}
\begin{figure*}[t]
    \centering
    \includegraphics[width=1.0\textwidth]{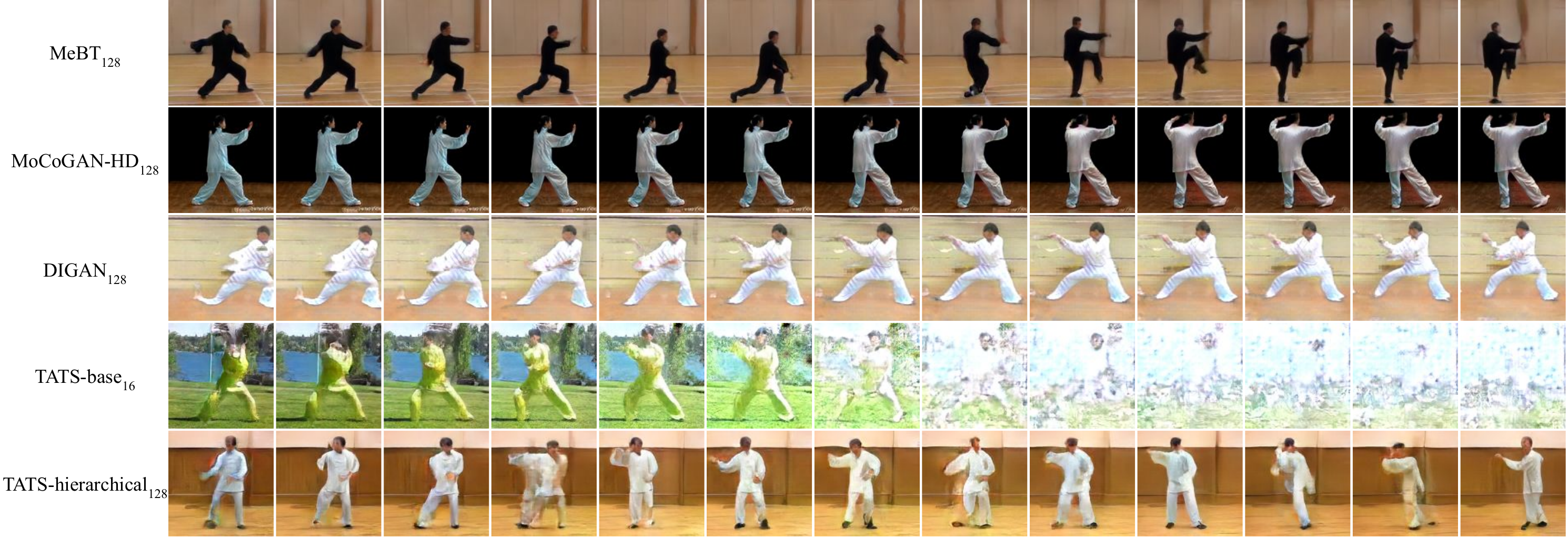}
    \caption{Generated videos on Taichi-HD. We displayed every 10th frame in the generated video. The subscript denotes the length of training videos. More samples can be found on the website.}
    \label{fig:supp_taichi}
\end{figure*}

The generated videos on Taichi-HD are shown in Figure \ref{fig:supp_taichi}. Unlike other baselines, MeBT could model the non-linear motions in Taichi by connecting the basic actions. Specifically, DIGAN showed difficulties in generating non-linear motions and MoCoGAN-HD could not keep the background and actor consistently. The frames generated by TATS-base become brighter as the video goes longer due to the error propagation, and TATS-hierarchical showed temporally ziggy motions by adopting two separately trained transformers. The ziggy motions of TATS-hierarchical are best viewed on the website.

\subsection{Qualitative Comparison on UCF-101}
\begin{figure*}[t]
    \centering
    \includegraphics[width=1.0\textwidth]{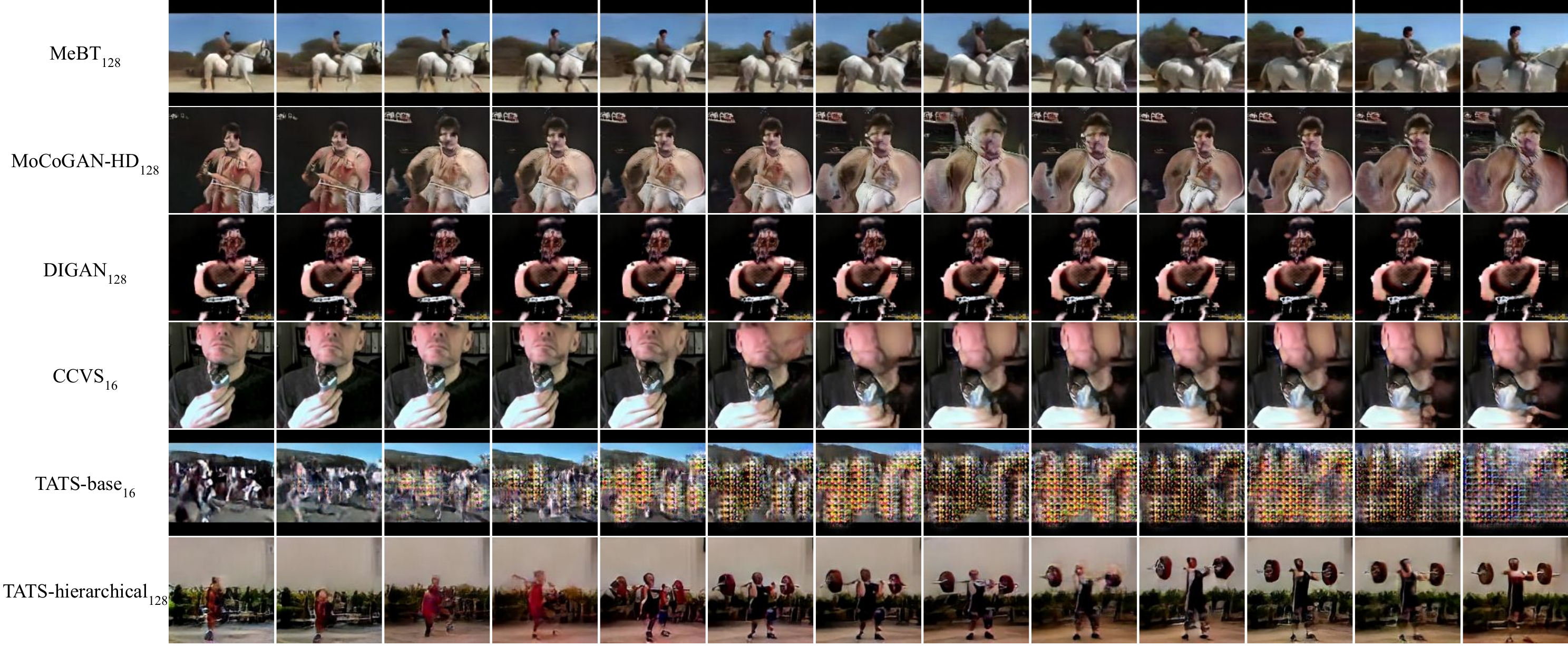}
    \caption{Generated videos on UCF-101. We displayed every 10th frame in the generated video. The subscript denotes the length of training videos. More samples can be found on the website.}
    \label{fig:supp_ucf}
\end{figure*}

The generated videos on UCF-101 are displayed in Figure \ref{fig:supp_ucf}. On the complex UCF-101 dataset, DIGAN and MoCoGAN-HD failed to model complex structures such as human faces. On the other hand, transformer-based models (TATS, MeBT) could model complex structures. However, TATS-base showed unnatural artifacts due to the error propagation, and TATS-hierarchical showed inconsistent quality between the keyframes generated by the hierarchical transformer and the interpolated frames. Compared to the baselines, MeBT generates complex motion that combines both camera motion and horse riding.



\begin{table*}[t]
    \centering
    \caption{Training configuration for all experiments. Subscripts denote the length of training videos.}
    \label{tab:train_config}
    \begin{tabular}{l|cccccc}
    \toprule
    Dataset & SkyTimelapse$_{16}$ & Taichi-HD$_{16}$ & UCF-101$_{16}$ & SkyTimelapse$_{128}$ & Taichi-HD$_{128}$& UCF-101$_{128}$\\
    \midrule
    Batch size & 24 & 128 & 128 & 40 & 48 & 48\\
    Learning rate & 1.08e-5 & 3e-5 & 3e-5 & 1.8e-5 & 3e-5 & 3e-5\\
    Training steps & 400k & 750k & 2.6M & 500k & 3M & 3.55M \\
    Dropout rate & 0.1 & 0 & 0 & 0.1 & 0 & 0 \\
    \bottomrule
    \end{tabular}
\end{table*}

\begin{table*}[t]
    \centering
    \caption{Decoding configuration for all experiments. Subscripts denote the length of training videos.}
    \label{tab:inference_config}
    \begin{tabular}{l|cccccc}
    \toprule
    Dataset & SkyTimelapse$_{16}$ & Taichi-HD$_{16}$ & UCF-101$_{16}$ & SkyTimelapse$_{128}$ & Taichi-HD$_{128}$& UCF-101$_{128}$\\
    \midrule
    $N_d$ & 0 & 0 & 0 & 64 & 64 & 64\\
    \hline
    $S$ & 32 & 64 & 128 & 32 & 32 & 32\\
    top-k & - & - & - & 32 & 32 & 32 \\
    $\gamma$ & cosine & cosine & cosine & cosine & cosine & cosine \\
    $\tau$ & 8 & 2 & 6 & 4 & 4 & 2 \\
    \hline
    $R$  & 2 & 2 & 4 & 2 & 2 & 32 \\
    temperature  & 0.7 & 0.3 & 0.7 & 0.7 & 0.1 & 0.1 \\
    $N_R$  & 2 & 8 & 4 & 2 & 4 & 2 \\
    \bottomrule
    \end{tabular}
\end{table*}


\end{document}